%% file: main.tex
\documentclass[twoside,twocolumn,10pt]{article}

\usepackage{wscg}           
\RequirePackage{ifpdf}
\ifpdf
 \RequirePackage[pdftex]{graphicx}
 \RequirePackage[pdftex]{color}
\else
 \RequirePackage[dvips,draft]{graphicx}
 \RequirePackage[dvips]{color}
\fi
\usepackage{tabularx}      

\ifdefined\review \else
\usepackage{nopageno}       
\fi

\usepackage{enumitem}
\usepackage{amsmath}
\usepackage{amssymb}
\usepackage{subfigure}
\usepackage[ruled,vlined]{algorithm2e}
\usepackage{textcomp}
\usepackage{todonotes}

\RequirePackage{lineno}
\usepackage{fancyhdr}

\DeclareMathOperator\card{card}

\title{A Flexible Pipeline for the Optimization of CSG Trees}

\ifdefined\review
\author{
	\parbox{0.25\textwidth}{\centering
		First Author\\[1mm]
		author's affiliation\\
		1st line of address\\
		Country (ZIP) code, City, State\\[1mm]
		e@mail
	}
	\hspace{0.05\textwidth}
	\parbox{0.25\textwidth}{\centering
		Second Author\\[1mm]
		author's affiliation\\
		1st line of address\\
		Country (ZIP) code, City, State\\[1mm]
		e@mail
	}
	\hspace{0.05\textwidth}
	\parbox{0.25\textwidth}{\centering
		Third Author\\[1mm]
		author's affiliation\\
		1st line of address\\
		Country (ZIP) code, City, State\\[1mm]
		e@mail
	}
}
\else 
\author{
	\parbox{0.5\textwidth}{\centering
		Markus Friedrich, Christoph Roch, Sebastian Feld, Carsten Hahn \\[1mm]
		Institute for Computer Science\\
		LMU Munich\\
		Oettingenstr. 67\\
		80538 Munich, Germany\\[1mm]
		\{first name.last name\}@ifi.lmu.de
	}
	\hspace{0.0\textwidth}
	\parbox{0.50\textwidth}{\centering
		Pierre-Alain Fayolle\\[1mm]
		Division of Information and Systems\\
		The University of Aizu\\
		Aizu-Wakamatsu City \\
		965-8580 Fukushima, Japan\\[1mm]
		fayolle@u-aizu.ac.jp
	}
}
\fi

\usepackage{url}
\makeatletter
\def\Uslash{\mathbin{\mathchar`\/}\@ifnextchar{/}{\kern-.15em}{}}
\g@addto@macro\UrlSpecials{\do \/ {\Uslash}}
\def\Ucolon{\mathbin{\mathchar`:}\@ifnextchar{/}{\kern-.1em}{}}
\g@addto@macro\UrlSpecials{\do : {\Ucolon}}
\makeatother

\begin{document}
	
\ifdefined\review
\pagenumbering{arabic}   
\setpagewiselinenumbers	
\modulolinenumbers[5]
\linenumbers
\setlength\linenumbersep{5pt}	
\fi

\setlength{\abovedisplayskip}{6pt}
\setlength{\belowdisplayskip}{6pt}
\setlist[itemize]{itemsep=1pt, topsep=0pt}

\twocolumn[{\csname @twocolumnfalse\endcsname

\maketitle  

\begin{abstract}
\noindent
CSG trees are an intuitive, yet powerful technique for the representation of geometry using a combination of Boolean set-operations and geometric primitives. 
In general, there exists an infinite number of trees all describing the same 3D solid.
However, some trees are optimal regarding the number of used operations, their shape or other attributes, like their suitability for intuitive, human-controlled editing. 
In this paper, we present a systematic comparison of newly developed and existing tree optimization methods and propose a flexible processing pipeline 
with a focus on tree editability.
The pipeline uses a redundancy removal and decomposition stage for complexity reduction and different (meta-)heuristics for remaining tree optimization.  
We also introduce a new quantitative measure for CSG tree editability and show how it can be used as a constraint in the optimization process. 

\end{abstract}

\subsection*{Keywords}
Evolutionary Algorithms, Quantum Annealing, Geometry Processing, CAD, CSG, Combinatorial Optimization

\vspace*{1.0\baselineskip}
}]


\copyrightspace

\input{sections/introduction}
\input{sections/relatedwork}
\input{sections/background}
\input{sections/problemstatement}
\input{sections/concept}
\input{sections/evaluation}
\input{sections/conclusion}

%
%


\bibliographystyle{alpha} 
\bibliography{main} 

\input{sections/evaluationimages}

%

\end{document}

%% file: sections/introduction.tex
\section{Introduction}
Constructive Solid Geometry (CSG) trees are a powerful representation scheme for 3D geometry and an important building-block of 3D modelling software \cite{requicha1980acm}. 
While the creation, reconstruction and conversion from other representations of CSG trees has been covered in the literature, few works deal with the optimization of a given tree.
\\
The nature of CSG tree expressions as a combination of geometric primitives (so-called halfspaces) with Boolean set-operations (union, intersection, complement) suggests a deeper investigation of related methods from \textit{switching function minimization}. 
This paper investigates an adaption of already proposed, as well, as newly developed methods for the CSG tree optimization problem and extends them to a robust and flexible pipeline. 
Furthermore, it introduces the idea that CSG tree optimization does not have to be restricted to the reduction of the tree size but should include improvements to the tree's \textit{editability}.
\\
We consider our CSG tree optimization pipeline as a solution to the following problems:
Given a hand-modeled CSG tree with sufficient complexity, users have difficulty keeping track of potentially redundant parts.
An automatic, but manually triggered, optimization procedure comes in handy.
Furthermore, automatic CSG tree reconstruction methods \cite{fayolle2016evolutionary,wu2018constructing} might result in trees that are not optimal in size and hard to edit manually. Thus, our method can be beneficial in this scenario as well.
\\
This paper makes the following contributions: 
\begin{itemize}
    \item The description of a comprehensive pipeline for optimizing the \textit{editability} of a CSG expression, 
    \item A novel, sampling-based tree size optimization procedure suitable for Quantum Annealing hardware,
    \item A recursively defined measure of spatial subtree proximity as an indicator for CSG tree \textit{editability},
    \item A multi-objective optimization using a Genetic Algorithm that aims to minimize the CSG expression size and maximize its proximity value in order to improve tree \textit{editability}.
\end{itemize}
The paper is organized as follows:
Section~\ref{ch:background} gives basic definitions and concepts. Section~\ref{ch:relw} provides references to related works, whereas Section~\ref{ch:prsta} defines the problem to solve. Our approach is described in Section~\ref{ch:concept} and evaluated in Section~\ref{ch:eval}. Section~\ref{ch:conclusion} concludes the paper.

%% file: sections/relatedwork.tex
\section{Related Work}
\label{ch:relw}
Construction of CSG trees from the Boundary Representation (B-Rep) of a solid was considered by Shapiro and Vossler in \cite{Shapiro1991,shapiro1991efficient,shapiro1993separation}. The approach is based on: identifying a set of halfspaces sufficient for representing the input solid, building a CSG expression by considering all products of halfspaces, or their complement, that are inside the input solid (so-called fundamental products), and minimizing this expression. In particular, \cite{Shapiro1991} considers different approaches for the minimization of two-level expressions, either coming from switching theory \cite{quine1952,mccluskey1956,orourke1982}, or based on geometric considerations.
\\
Related to these works, Buchele and Crawford \cite{buchele2004three} propose an algorithm for producing a CSG expression from the Boundary Representation of a solid by considering early factoring of dominant halfspaces. Such a factoring should help in limiting the size of the produced CSG expression. 
Andrews proposes to simplify a CSG expression (obtained from a B-Rep) by removing from each fundamental product cell, (spatially) distant primitives \cite{andrews2013user}. The reason is that distant primitives may be viewed by the user as unrelated to a given fundamental product cell, and thus could lead to unintuitive results. \\
Recently, works on reconstructing a CSG expression from a point-cloud have become popular, 
see for example \cite{fayolle2016evolutionary,wu2018constructing}
, among others. 
The approach described in \cite{fayolle2016evolutionary} tries to minimize the size of the CSG expressions produced by a GA by penalizing large expressions in the objective function. 
The method proposed in \cite{friedrich2019gecco} is using multi-objective optimization to prevent the growth of the generated CSG expressions. It also uses a decomposition scheme that prevents spatially distant primitives to be used in unrelated CSG sub-expressions.  
\\
Other works related to the optimization, or manipulation, of CSG expressions, such as \cite{rossignac2011tvcg}, try to improve the rendering time of the model, and do not necessarily help in minimizing the size of the expression or improving its editability. 

%% file: sections/background.tex
\section{Background}
\label{ch:background}
\subsection{CSG Tree Representation}
\subsubsection{Formal Representation}
We follow the formal definition of CSG trees from Shapiro et al. \cite{Shapiro1991}:
Given a solid's point-set $S$, its boundary $\partial S$ consists of patches of halfspaces $H_S$. Halfspaces are regular sub-sets of the universal point-set $W$ usually described by signed distance functions (SDF) $F_H$: $\{\mathrm{x} \in \mathbb{R}^3 : F_H(\mathrm{x})=0\}$.
A CSG tree expression $\Phi$ (in the following, upper-case Greek letters are used for CSG tree expressions) consists of halfspace literals $\{h_0,h_1,...\}$ and symbols for regularized set-operations $\{\cup^{*}, \cap^{*}, \setminus^{*}, -^{*}\}$. 
Applying $\Phi$ to the set of halfspaces $H_S$ results in a CSG representation $\Phi(H_S)$ of $S$ iff $|\Phi(H_s)| = S$, where $|\cdot|$ denotes the point-set induced by a CSG representation.
\\
A CSG representation of $S$ is in disjunctive normal form (DNF) if it contains a sum ($\cup^{*}$) of halfspace products ($\cap^{*}$-combined halfspaces or negated halfspaces, so-called implicants).
If each implicant of a DNF expression contains all halfspaces (or their negations), it is in a so-called disjunctive canonical form (DCF). In that case, an implicant is called canonical intersection term (CIT) or fundamental product (FP).
For example, the universal set $W$ can be decomposed in $2^n$ CITs in case of $n$ halfspaces being used.
\\
An implicant $\Psi$ is a so-called prime implicant of $S$ if $|\Psi| \in S $ and the removal of a single halfspace from $\Psi$ results in $|\Psi| \notin S$.
A dominant halfspace (DH) $g \in H_S$ is a halfspace for which $S = g \cup^{*} S$ is always true ($g$ is then also a prime implicant of $S$).
\\
Where necessary, we use the following abbreviations for CSG expressions (halfspaces and set-operations): $|h_0| \cup^{*} |h_1| := h_0 + h_1$, $|h_0| \cap^{*} |h_1| := h_0 \cdot h_1$, $\setminus^{*} |h_0| := \overline{h_0}$ and $|h_0| -^{*} |h_1| := h_0 - h_1$.

\subsubsection{SDF-based implementation}
In order to compute the point-set of a CSG representation (e.g., $|\Phi(H_s)|$) we use an SDF-based approximation with $\min$- and $\max$-functions \cite{ricci197constgeo,pasko1995function,shapiro2007semi}: 
\begin{itemize}
	\item Intersection: $|\Phi| \cap^{*} |\Psi| := \max(F_{\Phi}, F_{\Psi})$
	\item Union: $|\Phi| \cup^{*} |\Psi| := \min(F_{\Phi}, F_{\Psi})$
	\item Complement: $\setminus^{*} |\Phi| := -F_{\Phi}$ 
	\item Difference: $|\Phi| -^{*} |\Psi| := \max(F_{\Phi}, -F_{\Psi}) $
\end{itemize}
Here, $F_{\Phi}$ and $F_{\Psi}$ are the SDFs corresponding to expression $\Phi$ and $\Psi$.
We assume, as a convention, that $F<0$ in the interior of the corresponding solid $S$. 
\\
Note that $\min$- and $\max$-functions are not regularized set-operations in the strict sense \cite{shapiro1999well} but a sufficient approximation for our purposes.

\subsection{Metaheuristics for Combinatorial Problems}
\subsubsection{Genetic Algorithms}
Genetic Algorithms (GA) are metaheuristics for solving discrete or continuous optimization problems. 
The process is inspired by biology and consists of evolving a population of creatures. Each creature represents a candidate solution to the problem. 
Starting from a randomly initialized population, a GA produces an updated population at each iteration by: a) ranking creatures according to a fitness function, b) generating new creatures by mutating a selected creature from the previous population, or by combining a selected pair of creatures from the previous population, c) selecting a few creatures to be preserved into the next population (elitism). 
Selection is performed based on the rank of each creature. 
The process is iteratively repeated until a termination criterion is met. 

\subsubsection{Quantum Annealing}
Quantum Annealing (QA) is another metaheuristic for solving (in general, discrete) optimization problems. 
It is based on quantum physics to find low energy states of a system corresponding to the optimal solution of a problem. The QA algorithm is described by a time-dependent Hamiltonian $\mathcal{H}(t)$:
\begin{equation*}
\mathcal{H}(t)=s(t)\mathcal{H}_I+(1-s(t))\mathcal{H}_P
\end{equation*}
The QA process starts in the lowest-energy state of a so-called initial Hamiltonian $\mathcal{H}_I$. During the annealing process, the problem Hamiltonian $\mathcal{H}_P$ is introduced and the influence of the initial Hamiltonian is reduced (described by $s(t)$, which decreases from $1$ to $0$). At the end of the annealing process, one ends up in an eigenstate of the problem Hamiltonian, which actually encodes the objective function of the problem. If this transition is executed sufficiently slowly, the probability to find the lowest energy state of the problem Hamiltonian is close to $1$, w.r.t the adiabatic theorem \cite{albash2018adiabatic}. 
\\
To perform QA on D-Wave Systems Quantum Annealing hardware, one needs to encode the problem ($\mathcal{H}_P$) in a so-called Quadratic Unconstrained Binary Optimization (QUBO) problem, which is a unifying model for representing a wide range of combinatorial optimization problems. The functional form of the QUBO the quantum annealer is designed to minimize is:
\begin{equation}\label{eq:qubo}
\text{min } \mathrm{x}^TQ\mathrm{x} \quad \text{with }\mathrm{x} \in \{0,1\}^{n_Q}, 
\end{equation}
where $\mathrm{x}$ is a vector of binary variables of size $n_Q$, and $Q$ is an $n_Q \times n_Q$ real-valued matrix describing the relationship between the variables. Given the matrix $Q$, the annealing process tries to find binary variable assignments to minimize the objective function (Eq.~\ref{eq:qubo}). 

%% file: sections/problemstatement.tex
\section{Problem Statement}
\label{ch:prsta}
We focus on the optimization of a CSG tree's \textit{editability}:
Given a solid's point-set $S$, a halfspace set $H_S$ and a CSG tree expression $\Phi$ with $|\Phi(H_S)|=S$, find the CSG tree expression $\Phi_{opt}$ with the best \textit{editability} which is assumed to be determined by two quantitative metrics:
\begin{itemize}
    \item \textbf{Size:} The amount of literals and operations in $\Phi_{opt}$.
    \item \textbf{Proximity:} The ratio between the number of operations of $\Phi_{opt}$ whose operands imply point-sets that overlap to the number of operations with operands that imply disjoint point-sets.
    This property is defined recursively as follows:
    Given a node $\Psi$ of $\Phi_{opt}$, either $\Psi$ is a leaf (an halfspace) or it has two children (operands) $\Psi_1$ and $\Psi_2$ such that the implied solids intersect (that is: $|\Psi_1| \cap^* |\Psi_2| \neq \varnothing$).
\end{itemize}
A size-optimal tree has no redundant operands which makes tree modification easier and a tree with a high degree of proximity leads to more predictable behavior when sub-trees are transformed spatially. 

%% file: sections/concept.tex
\section{Concept}
\label{ch:concept}
\begin{figure}[!htbp]
	\centering
	\includegraphics[width=0.9\columnwidth]{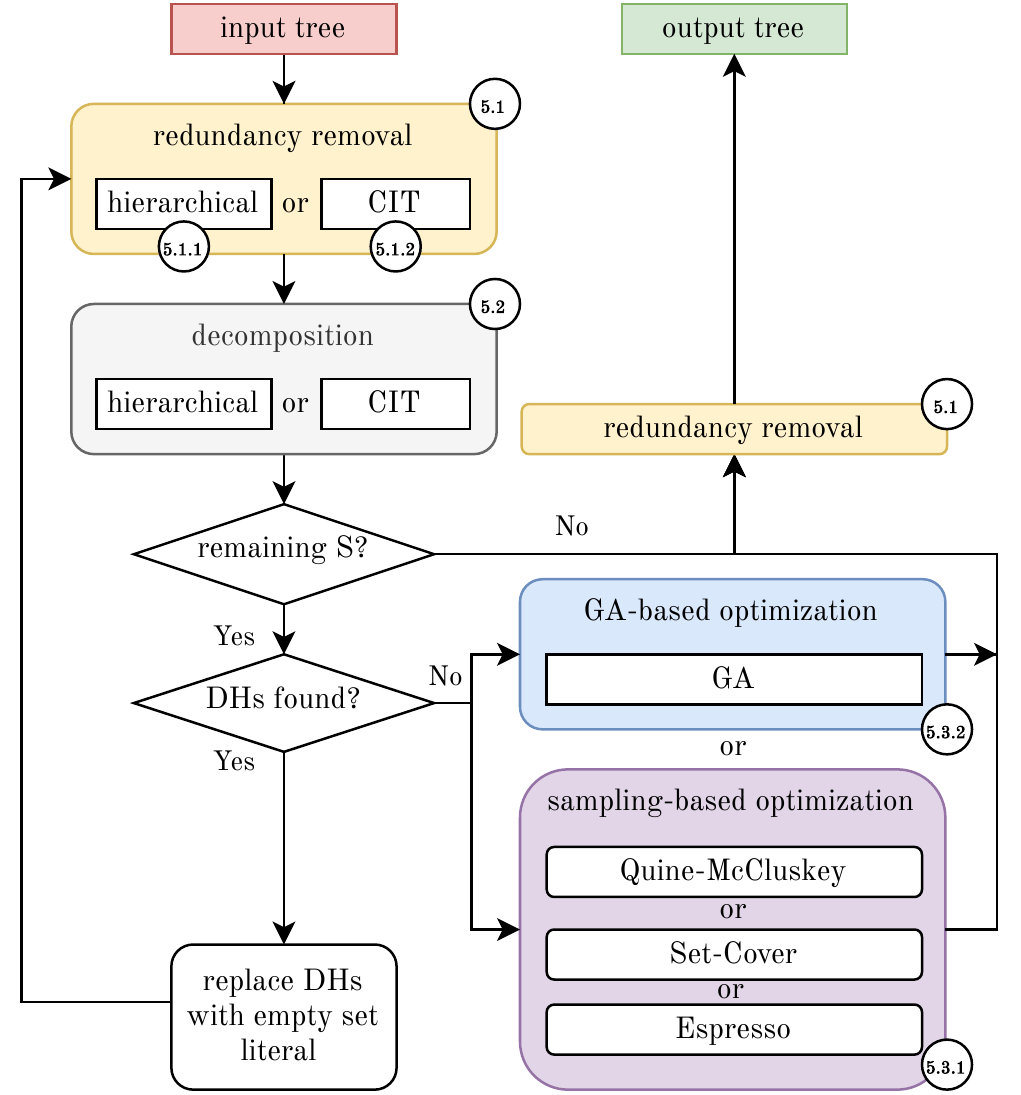}
	\caption{The proposed CSG tree optimization pipeline.} 
	\label{fig:csg_opt_pipeline}
\end{figure}

The CSG tree optimization process is 
depicted in Fig.~\ref{fig:csg_opt_pipeline}: 
First, redundant sub-expressions are removed (orange, Section \ref{ch:redrem}).
Then, a recursive decomposition scheme is applied that further shrinks the expression  
size (grey, Section \ref{ch:decomp}).
If an unoptimized expression (solid) remains, it is optimized with a separate optimization method (blue or purple, Section \ref{ch:rso}) which results -- after another run of the redundancy removal method -- in the final optimized CSG tree.
\subsection{Redundancy Removal}
\label{ch:redrem}
Our redundancy removal approach is inspired by a method described in \cite{Tilove1984}.
It uses the spatial information additionally given by the halfspaces used in the CSG tree (e.g., a sphere's location and radius):
\begin{itemize}
    \item If the sets described by the operands (halfspaces or subtrees) of an intersection operation do not have elements in common, i.e., the operand sets do not spatially overlap, the expression is replaced with the empty set expression $\varnothing$. 
    \item If the sets described by the operands of a union operation are identical, the expression is replaced with one of the operand expressions.
\end{itemize}
Empty set $\varnothing$ as well as universal set $W$ expressions are then replaced based on the following rules:
\begin{itemize}
    \item If one operand of an intersection expression is the empty set $\varnothing$, the expression is replaced with the empty set.
    \item If one operand of an intersection expression is the universal set $W$, the expression is replaced with the other operand.
    \item If one operand of a union expression is the empty set $\varnothing$, the expression is replaced with the other operand.
    \item If one operand of a union expression is the universal set $W$, the expression is replaced with $W$.
\end{itemize}
In addition, the complement of a complement operation is replaced with the operand of the inner complement operation.
The redundancy removal algorithm continuously iterates over the whole CSG tree until no rule applies anymore to the current result. \\
Especially relevant for the algorithm is the fast and robust evaluation of the empty set and identical set decision algorithms.
Since the identical set decision can be expressed as an empty set decision ($|\Phi| = |\Psi|  \Longleftrightarrow |\Phi| \cap^{*}(\setminus^{*}|\Psi|) = \varnothing \wedge |\Psi| \cap^{*}(\setminus^{*}|\Phi|) = \varnothing$ ), an empty set decision algorithm is sufficient.
\\
We use a sampling-based approach:
If the tree represents an empty set, its SDF value is positive (\textit{outside}) in the complete sampling domain. 
Two different sampling strategies are proposed: hierarchical sampling and CIT-based sampling.
\subsubsection{Hierarchical Sampling}
\label{ch:hiersa}
\begin{figure}[!htbp]
	\centering
	\includegraphics[width=0.7\columnwidth]{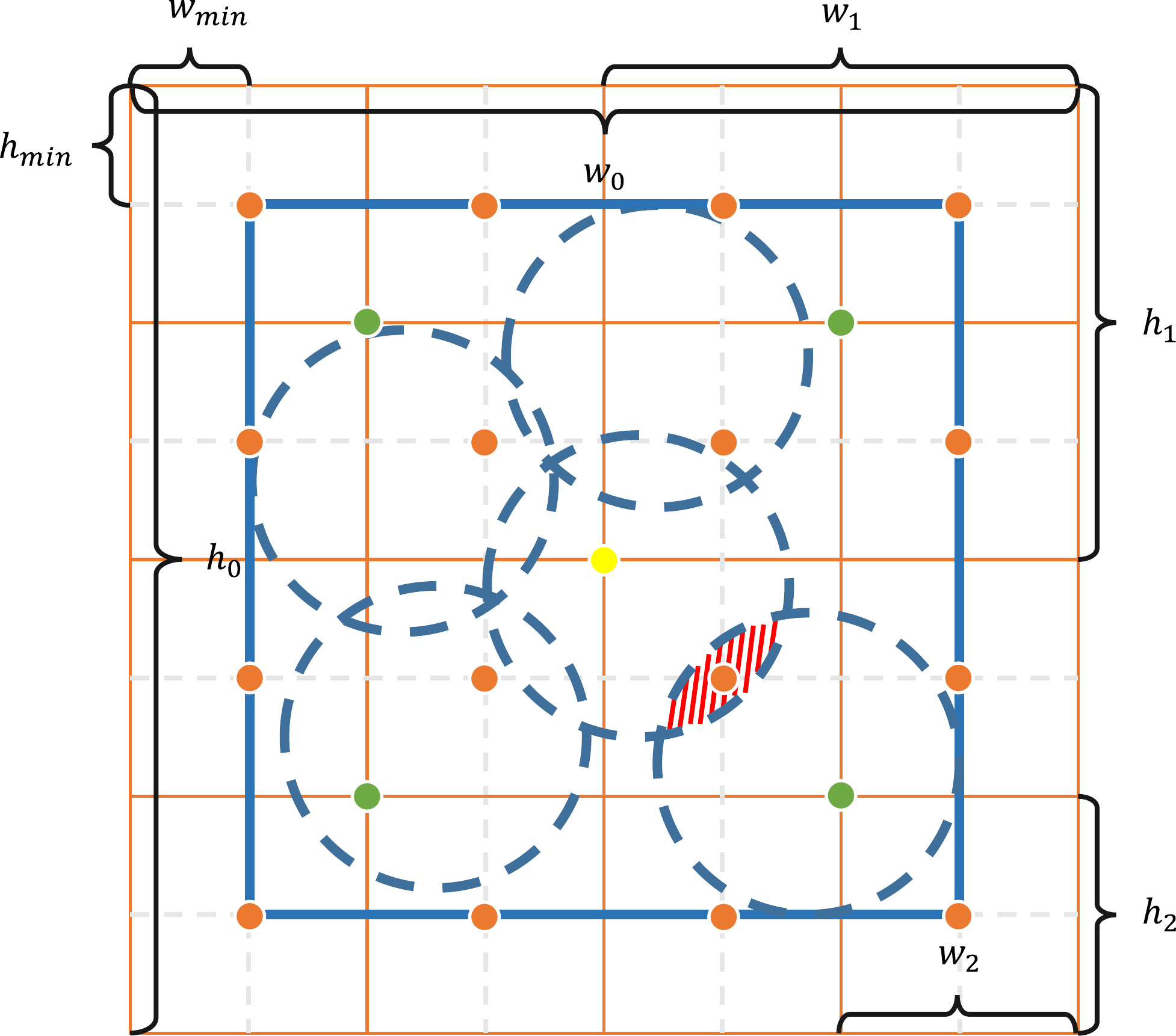}
	\caption{Hierarchical sampling strategy. The space defined by $(w_{0}, h_{0}, r_{0})$ is consecutively subdivided in sub-quadrants of sizes $s_i = (\frac{w_{i-1}}{2},\frac{h_{i-1}}{2},\frac{r_{i-1}}{2}), i=1,2,...$ until $w_i \leq w_{min} \vee h_i \leq h_{min} \vee r_i \leq r_{min}$. The center of each quadrant is the sampling position (yellow, green, orange). In the example, early stopping is possible since the red-hatched area, which marks the non-empty volume, is sampled at quadrant size $(w_{2}, h_{2}, r_{2})$ which is one level above the smallest possible quadrant size.} 
	\label{fig:red_opt_sampling}
\end{figure}

This sampling method is an Octree-based, hierarchical sampling of the CSG tree's SDF. 
The sampling point-set is defined by the width, height and depth of the axis-aligned bounding box (AABB) dimensions of the used halfspaces $(w_{0}, h_{0}, r_{0})$ and the user-defined minimum sampling cell size $(w_{min}, h_{min}, r_{min})$.
The coarse-to-fine hierarchical sampling methodology, as depicted in Fig.~\ref{fig:red_opt_sampling} allows for early stopping in case of a non-positive SDF value. 
To further speed-up the process, a lookup-table for already proven empty set expressions is used. 
\subsubsection{CIT-Based Sampling}
\label{ch:citsa}
This sampling method uses a sampling point-set consisting of a point for each CIT located inside the tree.  
CITs are retrieved with the method explained in Section \ref{ch:sbo}.
For each point we check if the tree's SDF is positive.
The approach can potentially be faster than hierarchical sampling since the used sampling point-set is usually smaller.
\subsection{Decomposition}
\label{ch:decomp}
The decomposition of a solid $S$ (that is described by a CSG tree $\Phi$ and halfspaces $H_S$) as proposed in \cite{Shapiro1991} is a tree expression that consists of a chain of halfspaces from $H_S$ that either dominate $S$ or $\overline{S}$ and a (potentially empty) remaining solid $S_{rem}$: 
\begin{equation}
\label{eq:dec}
S = |((...( S_{rem}\oplus d_1 ) \oplus ...) \oplus d_2) \oplus d_n|,
\end{equation}
where $\{d_1, ..., d_n\}$ is the set of dominating halfspaces and $\oplus$ is either $+$ if the following halfspace dominates $S$ or $-$ if it dominates $\overline{S}$. 
This decomposition is a size-optimal tree expression for $S$ since each halfspace appears only once \cite{Shapiro1991}. 
For example, in Fig.~\ref{fig:opt_sampling}, $h_0, h_2$ dominate $\overline{S}$ and $h_3$ dominates $S$, resulting in the decomposition $S=|((h_4-h_0)-h_2)+h_3|$.
Here, $S_{rem}$ contains $h_4$ and $h_1$ but since $h_1$ has no impact on the result, $S_{rem}$ equals $h_4$ and thus is replaced in the result expression.
Please note that we use the example in Fig.~\ref{fig:opt_sampling} differently than in Section \ref{ch:sbo}: The solid shown is considered $S$, which is decomposed to get $S_{rem}$, while in Section \ref{ch:sbo} the solid shown is considered $S_{rem}$ which is then further optimized.
\\
\textbf{Recursive Decomposition.}
If $S_{rem}$ is not empty, a size-optimal expression for it has to be found. 
Therefore, a (potentially not size-optimal) expression $\Phi_{rem}$ is computed by replacing all appearances of previously found dominating halfspaces in $\Phi$ with the empty set literal and applying the redundancy removal method (see Section \ref{ch:redrem}).
This is followed by another run of the decomposition technique.
This recursive process is continued until either the current $S_{rem}$ is empty, or no more dominating halfspaces can be found during decomposition. 
In the latter case, an optimal expression for $S_{rem}$ has to be found using different methods, see Section \ref{ch:rso}.
\\
\textbf{Sampling-Based Search for Dominant Halfspaces.} 
For the identification of the dominant halfspaces of $S$, we propose two different sampling strategies as already described in Section \ref{ch:redrem}: 
Firstly, a hierarchical sampling strategy similar to that proposed in Section \ref{ch:hiersa} can be used.
But instead of taking the whole AABB of $S$ as sampling volume, each halfspace $h$ in $H_S$ is separately tested using its corresponding AABB.
If all sampling points inside $|h|$ are inside $S$ as well, $h$ dominates $S$. 
If all sampling points inside $|h|$ are not elements of $S$, $h$ dominates $\overline{S}$. 
The early-stopping criteria is met if a sampling point is in $|h|$ but not in $S$.
Secondly, a CIT-based sampling strategy as discussed in Section \ref{ch:citsa} can be used.
\\
\textbf{Improving Proximity.}
Although decomposition can result in size-optimal expressions, the \textit{editability} criterion is also influenced by the proximity metric which is not considered by the proposed decomposition method.
To overcome this deficiency, we propose a simple spatial sorting scheme: 
The halfspaces in the chain of halfspaces (Eq.~\ref{eq:dec}) are arranged such that operands of $\oplus$ do always spatially overlap (if possible).



\subsection{Remaining Solid Optimization}
\label{ch:rso}
After decomposition, there might be a remaining solid $S_{rem} = |\Phi_{rem}(H_{rem})|$  left, for which an optimal expression has to be found. 
We investigate two optimization methods: 
Firstly, a sampling-based technique that generates tree expressions in DCF form that are then optimized using well-known two-level logic minimization techniques (Quine-McCluskey \cite{mccluskey1956}, Espresso logic minimizer \cite{brayton1984}) and a new approach based on a QUBO formulation of the Set Cover problem (Section \ref{ch:sbo}).
Secondly, a GA-based method is proposed that uses tree size and the proximity metric as part of its objective function (Section \ref{ch:ga}).
\subsubsection{Sampling-Based Optimization}
\label{ch:sbo}
\begin{figure}[!htbp]
	\centering
	\includegraphics[width=0.6\columnwidth]{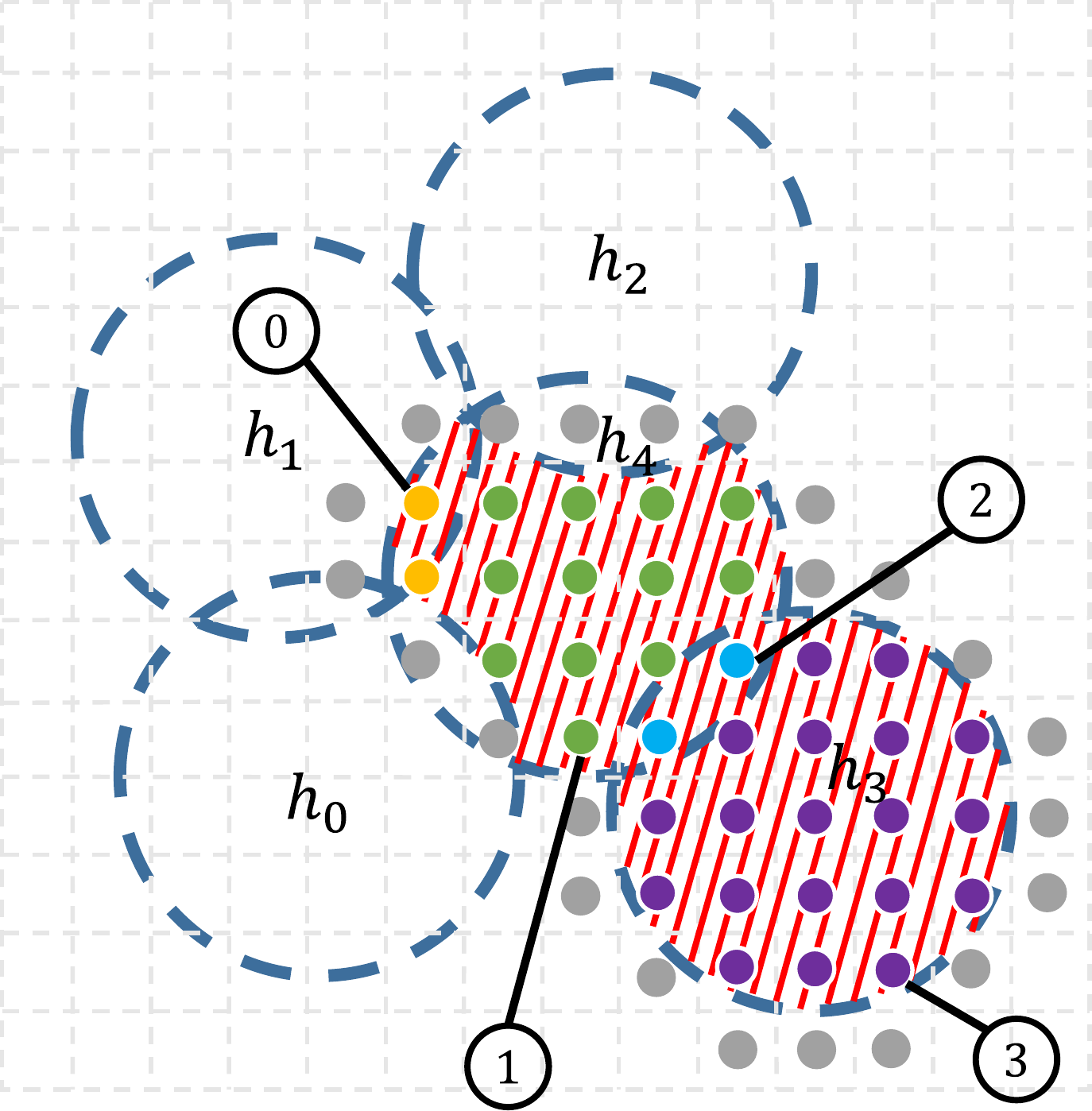}
	\caption{Example of the sampling step. The red-hatched area is the point-set to represent, $\{h_0,...,h_4\}$ is the halfspace set $H_{rem}$. The grey dots indicate sampling points outside the solid. The orange dots result in the implicant $\overline{h_0} \cdot h_1 \overline{\cdot h_2} \overline{\cdot h_3} \cdot h_4$ (0), the green dots in $\overline{h_0} \cdot \overline{h_1} \overline{\cdot h_2} \overline{\cdot h_3} \cdot h_4$ (1), the light blue dots in $\overline{h_0} \cdot \overline{h_1} \overline{\cdot h_2} \cdot h_3 \cdot h_4$ (2) and the purple dots in $\overline{h_0} \cdot \overline{h_1} \overline{\cdot h_2} \cdot h_3 \cdot \overline{h_4}$ (3).}
	\label{fig:opt_sampling}
\end{figure}

\textbf{Sampling.} The AABB of $S_{rem}$ is sampled. 
For each sample, the tree's SDF is evaluated to decide if the point is \textit{inside}.
If it is \textit{inside}, each halfspace SDF in $H_{rem}$ is evaluated as well.
If the sampling point is located \textit{inside} the halfspace, the halfspace is part of a CIT.
If not, then its complement is.
Finally, the CIT is added with a $\cup^{*}$ operation to the resulting DCF expression $\Phi_{DCF}$.
See Fig.~\ref{fig:opt_sampling} for an example.
\\
\textbf{DCF Minimization.} 
Besides the already mentioned classic methods for DNF minimization (Quine-McCluskey, Espresso logic minimizer), we propose a third option:
First, the prime implicants of $\Phi_{DCF}$, $P_{DCF}$, are computed by directly applying the definition of a prime implicant (see Section \ref{ch:background}) to each CIT in $\Phi_{DCF}$.
Using the example of Fig.~\ref{fig:opt_sampling}, the prime implicants would be $h_3$ and $\overline{h_0 \cdot h_2} \cdot h_4$ (Note that we do not consider decomposition for this example).
\\
The problem of finding all relevant prime implicants is then formulated as a Set Cover problem:
The set to cover, $U$, is the set of indices of all CITs located inside $|\Phi_{DCF}|$ (in Fig.~\ref{fig:opt_sampling}: $U=\{0,1,2,3\}$).
Each prime implicant in $P_{DCF}$ covers a subset of $U$ resulting in a collection of subsets $V$ with elements $V_{k} \subseteq U, 1 \leq k \leq \text{card}(P_{DCF})$ (in Fig.~\ref{fig:opt_sampling}: $V=\{h_3:\{2,3\},\overline{h_0 \cdot h_2} \cdot h_4:\{0,1,2\}\}$).
Within the Set Cover problem, one has to find the smallest possible number of subsets from $V$, such that their union is equal to $U$. 
This problem was proven to be NP-hard \cite{karp1972reducibility}. 
In \cite{lucas2014ising} the QUBO formulation for the Set Cover problem is given by: 
\begin{equation*}
\begin{aligned}
\mathcal{H}_A ={} & A \sum_{\alpha=1}^{\text{card}(U)}  \left(1- \sum_{m=1}^{\text{card}(P_{DCF})} x_{\alpha,m} \right)^2 + \\
      & A \sum_{\alpha=1}^{\text{card}(U)}  \left(\sum_{m=1}^{\text{card}(P_{DCF})}mx_{\alpha,m} -\sum_{k:\alpha \in V_k} x_{k} \right)^2, 
\end{aligned}
\end{equation*}
and
\begin{equation}\label{eq:QUBO SC-hard-2}
\mathcal{H}_B = B \sum_{k=1}^{\text{card}(P_{DCF})} x_k, 
\end{equation}
with $x_k$ being a binary variable which is $1$, if set $V_k$ is included within the selected sets, and $0$ otherwise.
$x_{\alpha,m}$ denotes a binary variable which is $1$ if the number of selected subsets $V_{k}$ which include element $\alpha$
is $m \geq 1$, and $0$ elseways. The first energy term in $\mathcal{H}_A$ imposes the constraints that for any given $\alpha$ exactly one
$x_{\alpha,m}$ must be $1$, since each element of $U$ must be included a fixed number of times. The second term in $\mathcal{H}_A$ states that the number
of times that we declared $\alpha$ was included is in fact equal to the number of subsets $V_k$ we have included with $\alpha$ as an element. $A$ is a penalty value, which is added on top of the solution energy, described by $\mathcal{H} = \mathcal{H}_A + \mathcal{H}_B$, if a constraint was not fulfilled, i.e., one of the two terms are unequal to $0$. Therefore adding a penalty value states a solution as invalid. Additionally, the Set Cover problem minimizes over the number of chosen subsets $V_{k}$, as stated in (Eq.~\ref{eq:QUBO SC-hard-2}).
For a given problem instance, $\mathcal{H}$ is transformed into the QUBO formulation required by QA hardware (Eq.~\ref{eq:qubo}) and minimized.
\\
Note that the prime implicant selection via Set Cover is not needed if $H_{rem}$ does not contain halfspaces that fully contain other halfspaces and no separating halfspaces \cite{Shapiro1991} are used.
Furthermore, the sampling step can be omitted if the input expression is already in DNF form.
\subsubsection{GA-Based Optimization}
\label{ch:ga}
The methods described in Section \ref{ch:sbo} are two-level minimization techniques, which in general do not result in size-optimal trees \cite{Shapiro1991}.
Moreover, other optimization goals like, for example, the proximity metric are not considered.
Their main advantages are a possible short execution time and the fact that non-optimality of sufficiently small trees (e.g., after decomposition as explained in Section \ref{ch:decomp}) has usually smaller and thus negligible negative effects on tree size and proximity.
\\
In order to find possibly better trees (w.r.t. size and proximity), the corresponding problem is formulated as a combinatorial optimization problem over all possible trees given a set of halfspaces $H_{rem}$ and set-operations.
Let $\Phi_{rem}$ be the input CSG tree expression that needs to be optimized. 
A GA is used to solve this optimization problem. 
Each creature $\Phi_c$ in the population of the GA represents a potential CSG tree expression. The same mutation, crossover and selection as in \cite{fayolle2016evolutionary} are used.
Additional details about the GA are provided below. 
\\
\textbf{Initialization.} 
The population is initialized with a mixture of randomly generated trees with a maximum size equal to the size of the input tree $\Phi_{rem}$ and copies of the input tree.
\\
\textbf{Pre-processing.} 
To reduce the computational effort of evaluating the fitness function in the GA, we compute a limited number of sample points. 
This is done by computing the CITs of $\Phi_{rem}$ via sampling as described in Section \ref{ch:sbo} with the difference that CITs that are located outside of $\Phi_{rem}$ are considered as well. 
During the sampling process, a point in each CIT is added to the sampling point-set $S_{in}$ if the corresponding CIT is inside of $\Phi_{rem}$. 
Otherwise, it is added to $S_{out}$.
\\
\textbf{Ranking.} The fitness of a creature $\Phi_c$ corresponding to a given CSG tree is given by
$$
f(\Phi_c)= \alpha \cdot f_{geo}(\Phi_c) + \beta \cdot f_{prox}(\Phi_c) + \gamma \cdot f_{size}(\Phi_c), 
$$
where $f_{geo}(), f_{prox}()$ and $f_{size}()$ are the geometric score, the proximity score and the size score, respectively, and $\alpha, \beta$ and $\gamma$ are user-defined parameters. 
The geometric score counts how many points from $S_{in}$ are elements of $|\Phi_c|$ and how many points from $S_{out}$ are not in $|\Phi_c|$:
$$
f_{geo}(\Phi_c) = f_{in}(\Phi_c) + f_{out}(\Phi_c),
$$
with
$$
f_{in}(\Phi_c) = \frac{1}{\card(S_{in})} \sum_{s \in S_{in}}
\begin{cases}
1, & \text{if } |F_{\Phi_c}(s)| \leq \epsilon_p \\
0, & \text{otherwise} 
\end{cases}
$$
and 
$$
f_{out}(\Phi_c) = \frac{1}{\card(S_{out})} \sum_{s \in S_{out}}
\begin{cases}
1, & \text{if } |F_{\Phi_c}(s)| > \epsilon_p \\
0, & \text{otherwise} 
\end{cases},
$$
where $\card(S)$ is the cardinality of the point-set $S$. 
\\
The proximity score tries to enforce that two operands of a Boolean operation are spatially connected.
It is an implementation of the proximity metric (Section~\ref{ch:prsta}):
$$
f_{prox}(\Phi) = \frac{P_{rec}(\Phi)}{\#|\Phi|},
$$
where $\#|\Phi|$ is the number of nodes in the tree corresponding to the creature $\Phi$ and the function 
$$
P_{rec}(\Phi) = 
\begin{cases}
1, \text{ if $\Phi$ is a leaf node} \\
P_{rec}(\Phi_1) + P_{rec}(\Phi_2) + \Delta(\Phi_1,\Phi_2) \; \text{otherw.}
\end{cases}, 
$$
where $\Phi_1$ and $\Phi_2$ are the two children of $\Phi$ (if it is not a leaf node), and 
$$
\Delta(\Phi_1,\Phi_2) = \begin{cases}
1 \text{ if } |\Phi_1| \cap^* |\Phi_2| \neq \varnothing \\
0 \text{ otherwise}
\end{cases}.
$$
\\
Finally, the size score tries to favor the simplest (shortest) possible CSG tree and corresponds to
$$
f_{size}(\Phi_c)=\frac{\#|\Phi_{rem}|-\#|\Phi_c|-f_{size}^{min}}{f_{size}^{max}-f_{size}^{min}}, 
$$
where $\#|\Phi_{rem}|$ and $\#|\Phi_c|$ are the number of nodes in the CSG trees $\Phi_{rem}$ and $\Phi_c$, $f_{size}^{min}$ is the minimum and $f_{size}^{max}$ the maximum tree size within the current iteration's population.
\\
\textbf{Termination.}
The GA terminates if either a user-defined maximum number of iterations is reached or the score does not improve over a number of iterations.

%% file: sections/evaluation.tex
\section{Evaluation}
\label{ch:eval}
\subsection{Data Acquisition}
We prepared $11$ different, hand-crafted CAD models for our experiments (see Fig.~\ref{fig:model}) with a complexity comparable to models commonly used in CSG tree reconstruction tasks \cite{fayolle2016evolutionary, du2018inversecsg}. 
To simulate particular levels of \textit{sub-optimality}, we have implemented a generator, which takes a CSG tree as input and iteratively adds redundant parts at random positions in the tree,  based on the following strategies: 
\\
\textbf{Copied Subtree Insertion (CSI):} Insert a union or intersection at a random position with both operands being copies of the subtree at that position.
\\
\textbf{Double Negation Insertion (DNI):} Insert two chained negations at a random position.
\\
\textbf{Distributive Law Insertion (DLI):} Apply the distributive laws $A(B+C)=(A \cdot B) + (A \cdot C) $ or $A+(B \cdot C) = (A + B) \cdot (A + C)$ to a random subtree.
\\
\textbf{Absorption Law Insertion (ALI):} Apply the absorption laws $A=A+( A\cdot B) $ or $A=A\cdot (A+B))$ to a random subtree.
\\
\textbf{GA-based Redundancy Insertion (GRI):} Use the GA (Section \ref{ch:ga}) with negative size and proximity weight to produce trees with redundant parts.
\\
A particular run of the generator can be parameterized by the tuple $(N_{\text{iter}}, P_{CSI}, P_{DNI}, P_{DLI},P_{ALI}, P_{GRI})$, where $N_{\text{iter}}$ is the number of iterations of the generator and $P_{...}$ are the probabilities of the insertion strategies.
These artificially introduced redundancies cover all sorts of redundancies potentially appearing in CSG trees. 
\begin{table*}[!htbp]
\label{tab:models}
\centering
\begin{tabular}{lll}
\hline
 Models (number of nodes, proximity score, (dimensions))                & Data set $1$            & Data set $2$            \\
\hline
 model1 (9, 0.75, (20.0$\times$22.0$\times$20.0))    & (97, 0.442)  & (78, 0.35)   \\
 model2 (13, 0.833, (15.8$\times$31.7$\times$11.3))  & (41, 0.952)  & (23, 0.727)  \\
 model3 (39, 0.474, (21.0$\times$6.0$\times$21.0))   & (79, 0.675)  & (57, 0.536)  \\
 model4 (27, 1, (13.4$\times$13.4$\times$12.0))      & (65, 0.97)   & (97, 0.708)  \\
 model5 (19, 0.667, (24.0$\times$18.0$\times$27.0))  & (59, 0.645)  & (48, 0.76)   \\
 model6 (19, 0.556, (23.1$\times$10.0$\times$10.0))  & (99, 0.667)  & (43, 0.857)  \\
 model7 (91, 0.706, (21.6$\times$7.4$\times$21.8))   & (144, 0.725) & (151, 0.728) \\
 model8 (73, 0.722, (31.5$\times$12.7$\times$4.5))   & (145, 0.74)  & (97, 0.813)  \\
 model9 (171, 0.471, (29.7$\times$3.84$\times$30.1)) & (191, 0.51)  & (216, 0.519) \\
 model10 (17, 0.875, (13.0$\times$12.5$\times$13.0)) & (41, 0.864)  & (32, 0.563)  \\
 model11 (37, 0.789, (26.0$\times$13.0$\times$22.0)) & (86, 0.894)  & (66, 0.8)    \\
\hline
\end{tabular}
\caption{Models and their inflated variants. For each data set, (number of nodes, proximity score) is depicted.}
\end{table*}
Table \ref{tab:models} lists properties of the input models and their generated inflated versions.
Data set 1 uses all possible inflation mechanisms with the same probability for $10$ iterations $(10,1.0,1.0,1.0,1.0,1.0)$, data set 2 uses GA inflation only for $20$ iterations $(20,0.0,0.0,0.0,0.0,1.0)$. 
All sampling techniques use a step size of $0.1$.

\subsection{Execution Times}
For evaluating the execution time, we used a quad core CPU @2.80 GHz and 16GB of RAM.
Set covers are computed without QA hardware acceleration.
\subsubsection{Pipeline}
For the pipeline steps \textit{Redundancy Removal} and \textit{Decomposition}, different options are possible: a) decomposition or b) the redundancy removal method can use the hierarchical sampling strategy (0) or CIT-based sampling (1) and c) redundancy removal can be used (1) or not (0).
Options a) - c) can be combined resulting in $6$ possible configurations.
A particular configuration is identified by a binary $3$-tuple, e.g., $(1,0,1)$ for decomposition with CIT-based sampling and redundancy removal with hierarchical sampling.
All cases use Espresso for the remaining solid optimizer.
Results are given in Fig.~\ref{fig:pipeline_timings_1} and \ref{fig:pipeline_timings_2}.
For both data sets, CIT-based sampling is inferior in any configuration due to its processing time. The initial redundancy removal does not have any positive impact on execution times as well (exception: model $3$ and model $10$ in both data sets).

\subsubsection{Remaining Solid Optimization}
Results for data set $1$ are given in Table \ref{tab:rso_timings}. 
\begin{table}[!htbp]
\label{tab:rso_timings}
\centering
\begin{tabular}{lllll}
\hline
 Models  & Q-McC & Espresso & Set Cover & GA \\
\hline
 model2  & 1572 & 1066 & 17305 & 21526 \\
 model8  & - & 1163 & 111017 & 79829 \\
 model10 & 3928 & 940 & 3347 & 22094  \\
 model11 & 3879 & 900 & 8594 & 90971 \\
 \hline
 \end{tabular}
\caption{RSO timings for data set $1$ in [ms].}
\end{table}
For data set $2$ results are similar and thus are not shown due to space constraints.
Overall the fastest method is Espresso, followed by the Quine-McCluskey method (exception: model $10$ where Set Cover is faster).
Within the Set Cover method, the prime implicant generation part is the most time expensive whereas solving the Set Cover problem is neglectable ($\sim 1$-$3$ms).   
The slowest method is the GA-based optimization with the exception of model $8$, where Set Cover is the slowest.

\subsection{Optimization Characteristics}
All experiments use the pipeline configuration $(0,0,1)$.
\subsubsection{Pipeline}
For both data sets, resulting size (Fig.~\ref{fig:opt_size}) and proximity (Fig.~\ref{fig:opt_prox}) for models $1$, $3$, $4$, $5$, $6$, $7$ and $9$ are the same for all methods since for these models, the remaining solids are empty after decomposition and thus no remaining solid optimization is necessary.
\\
For both data sets, the GA produces the best size results (exception: model $11$) but not always the best proximity results (e.g., data set $2$, model $2$).
This can be explained by two factors: The GA's objective function's size weight used in our experiments is greater than its proximity weight ($10$ vs. $1$) and larger trees have more redundancies and thus tend to have higher subtree overlap resulting in higher proximity scores.
This explains as well the results of the Set Cover method which are the worst in terms of tree size but among the best regarding proximity.
Compared to the original, hand-crafted input trees, resulting trees have equally good or better size and proximity values for both data sets (exception: size for model $8$, data set $2$).
Also worth mentioning is that the Quine-McCluskey method cannot handle the size of model $8$.
\subsubsection{GA Specifics}
For the GA, we used a population size of $150$, a mutation rate of $0.3$, a crossover rate of $0.4$ and a tournament selection $k=2$. 
Objective function weights are $\alpha=50, \beta=1, \gamma=10$. 
The maximum iteration count is $1000$, and after $500$ iteration without score change, the GA is terminated as well.
Fig.~\ref{fig:ga_models} shows one of the main advantages of the GA approach for remaining solid optimization: It is possible to manually select the best size/proximity trade-off solution for a particular use-case after the optimization process has finished.

%% file: sections/conclusion.tex
\section{Conclusion}
\label{ch:conclusion}
In this paper, we proposed a flexible new pipeline for the efficient optimization of a CSG tree's editability. 
We have evaluated different pipeline combinations with a representative set of models. 
\\
It is possible to use the sampling-based optimization method (Section \ref{ch:sbo}) in situations where the input expression is not known in advance (only the solid point- and halfspace-set).
In that case, another strategy for \textit{inside-outside} decisions would be needed. 
\\
Furthermore, the decomposition technique (Section \ref{ch:decomp}) can be also used for the task of CSG tree reconstruction from other geometry representations (e.g., point clouds or B-Rep representations). 
Therefore, the dominant halfspace detection must be adapted according to the geometry representation of the input and the discussed recursion would end after the first iteration without applied redundancy removal (which is not possible without a given input tree). A GA-based approach \cite{fayolle2016evolutionary} (or the aforementioned variant of the sampling-based optimization method) could be used to find a tree expression for the remaining solid. 

%% file: sections/evaluationimages.tex

\begin{figure*}[htb]
	\centering
	\subfigure[Model 1]{\includegraphics[width=25mm]{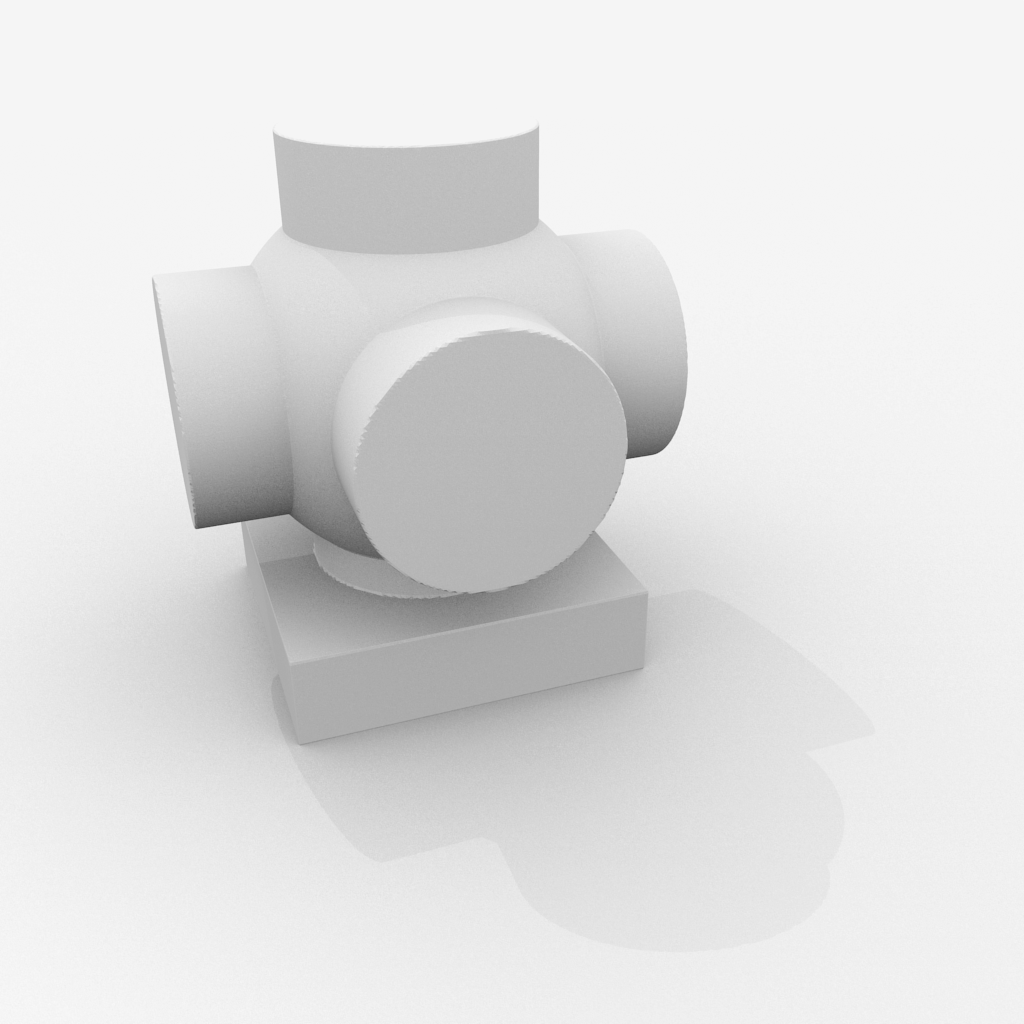}\label{fig:m1}}
	\subfigure[Model 2]{\includegraphics[width=25mm]{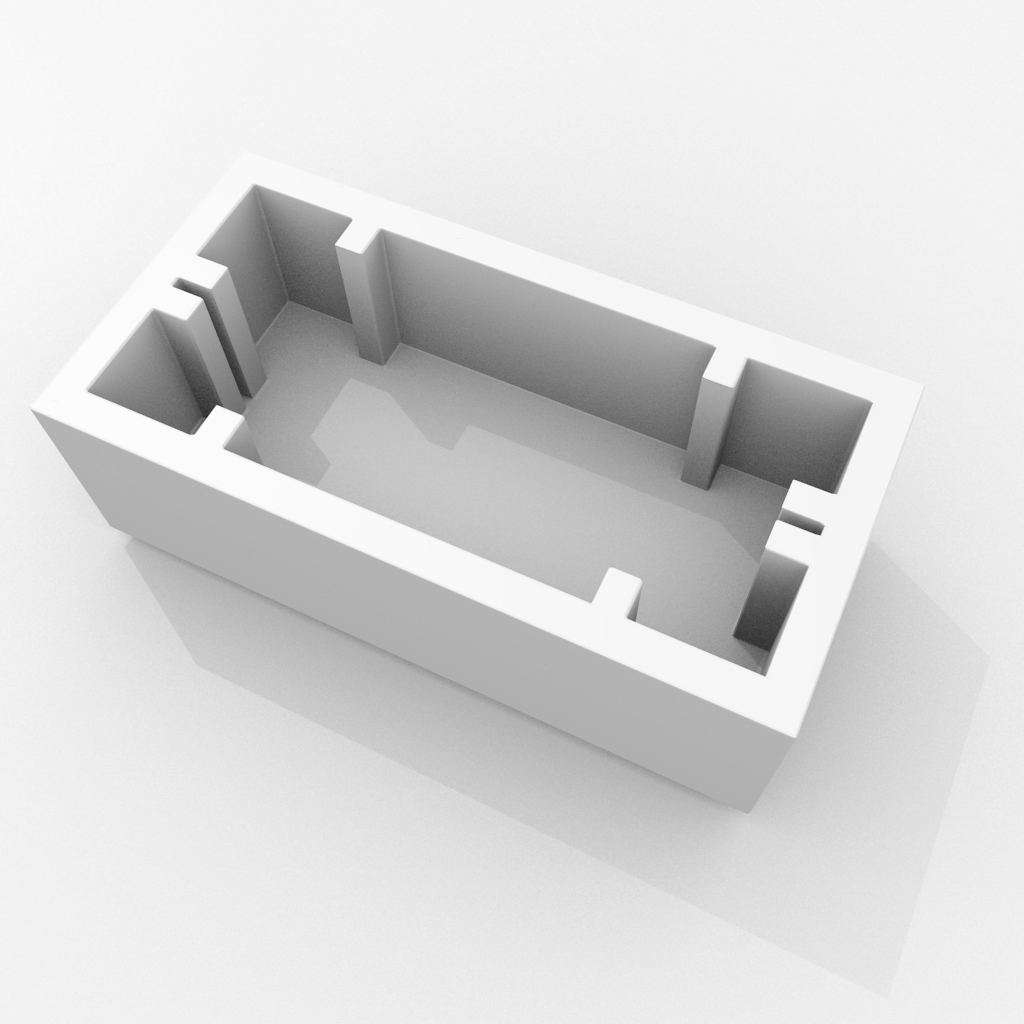}\label{fig:m2}}
	\subfigure[Model 3]{\includegraphics[width=25mm]{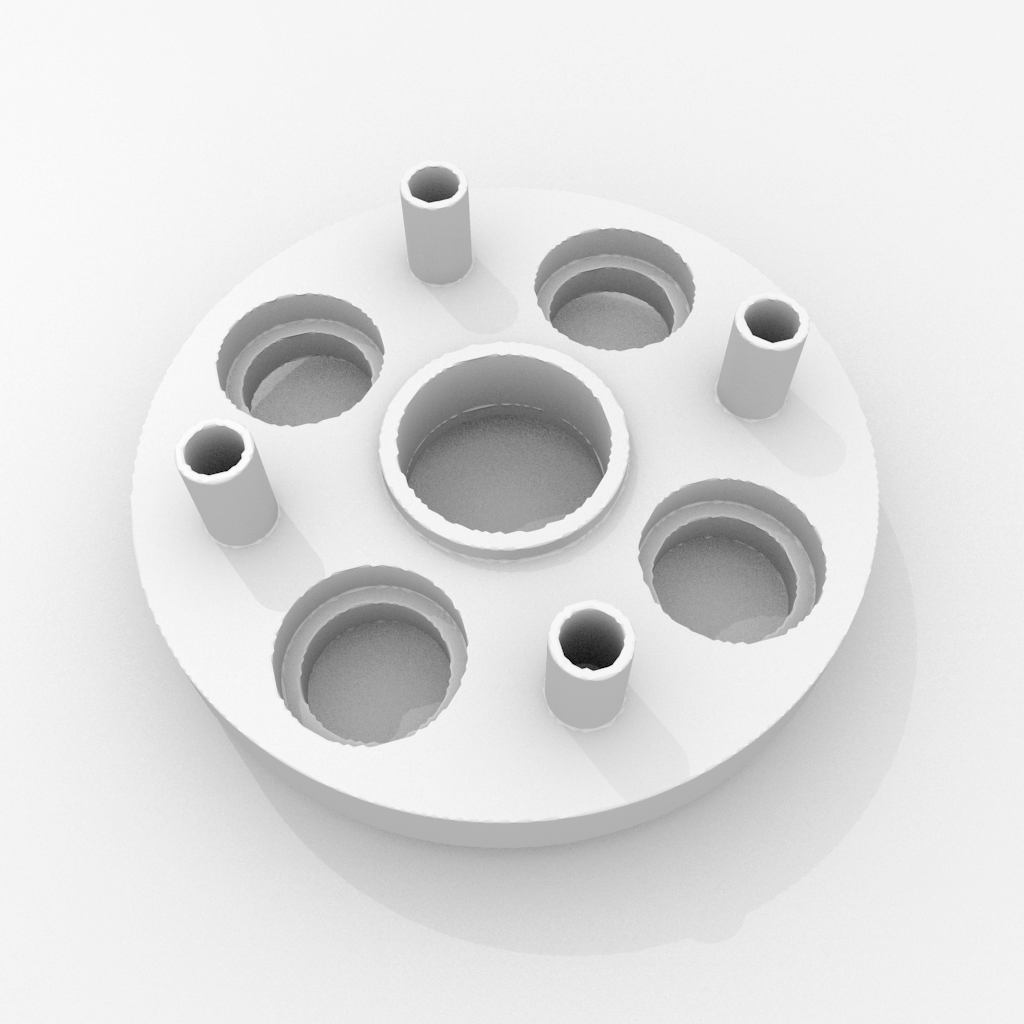}\label{fig:m3}}
	\subfigure[Model 4]{\includegraphics[width=25mm]{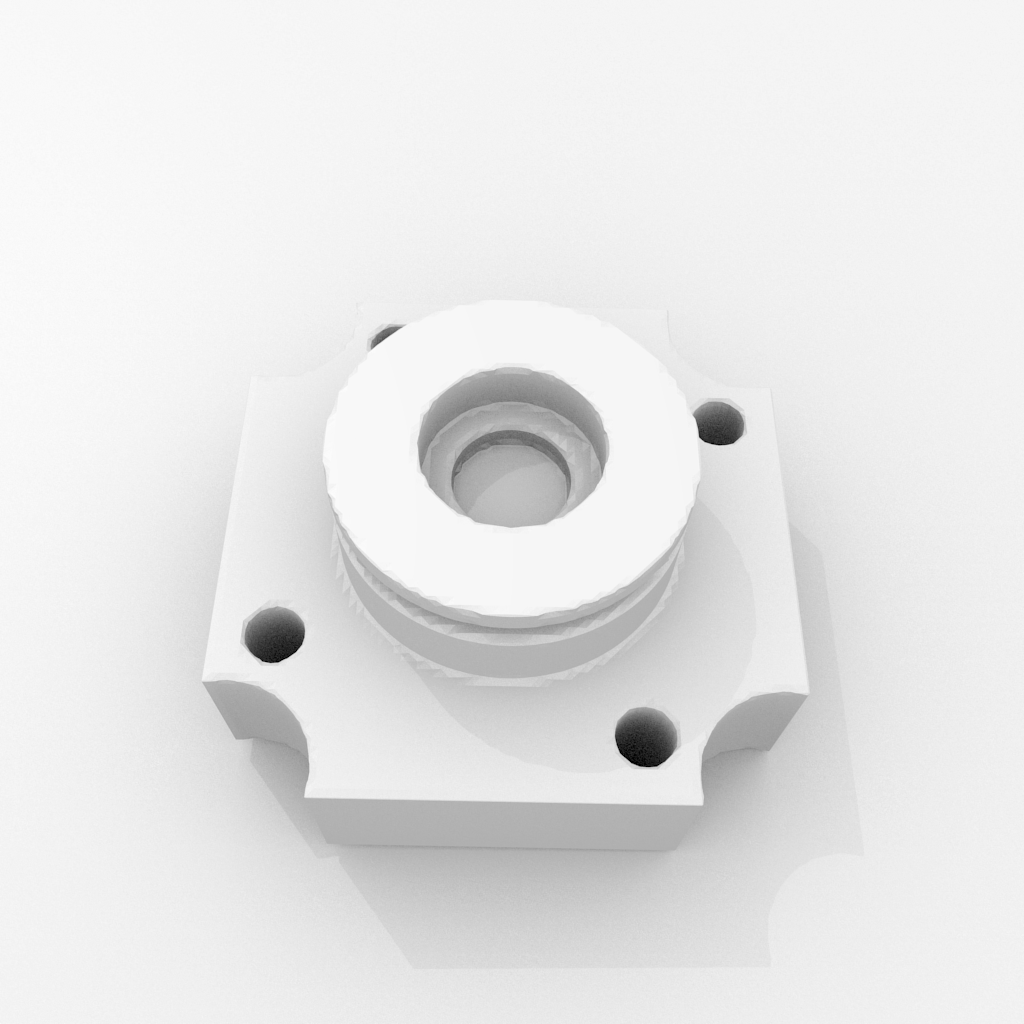}\label{fig:m4}}
	\subfigure[Model 5]{\includegraphics[width=25mm]{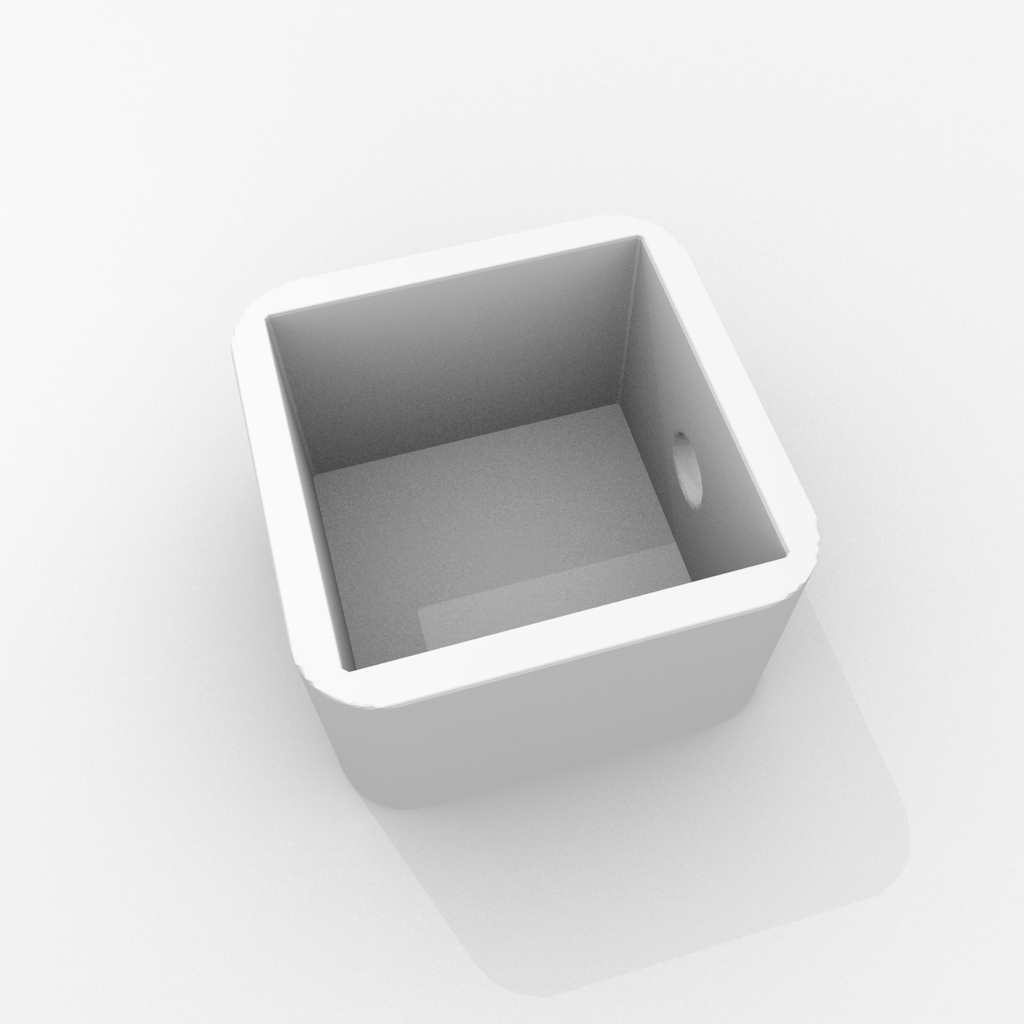}\label{fig:m5}}
	\subfigure[Model 6]{\includegraphics[width=25mm]{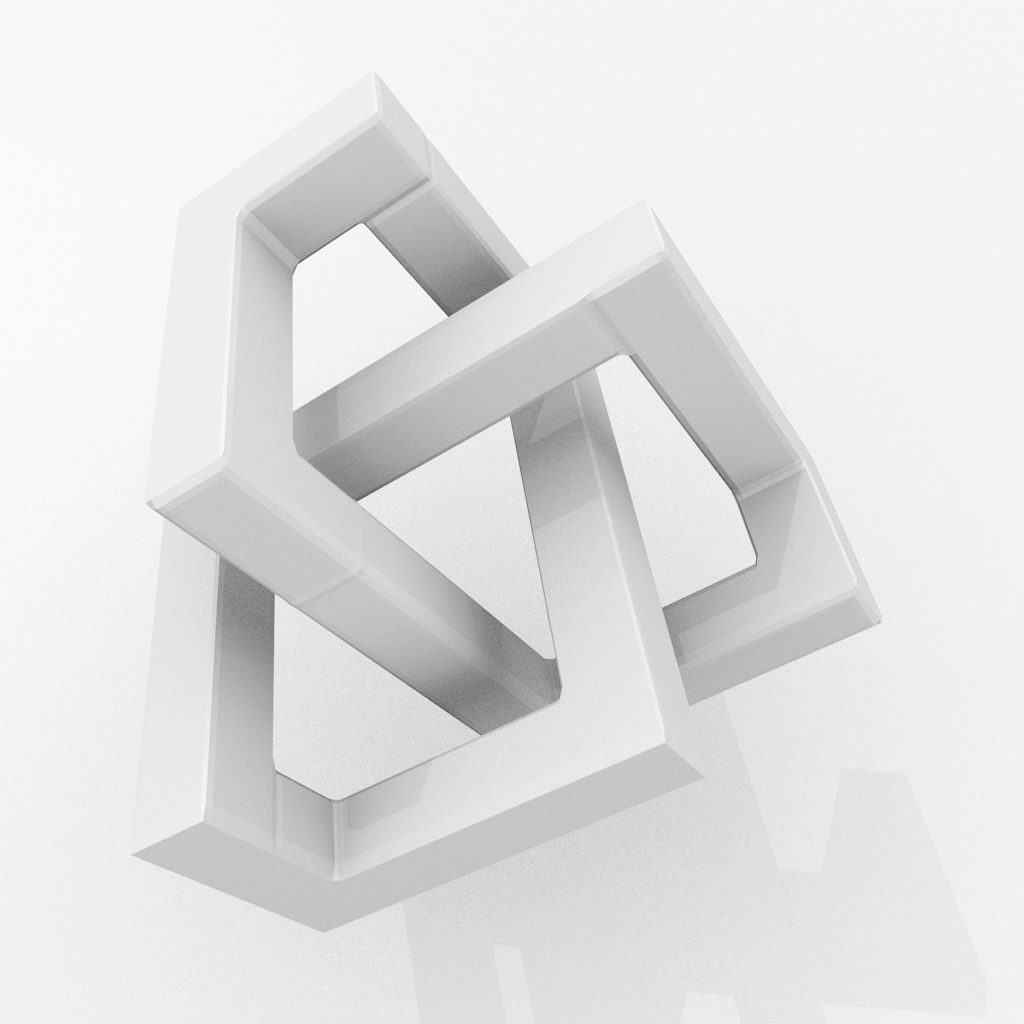}\label{fig:m6}}
	\subfigure[Model 7]{\includegraphics[width=25mm]{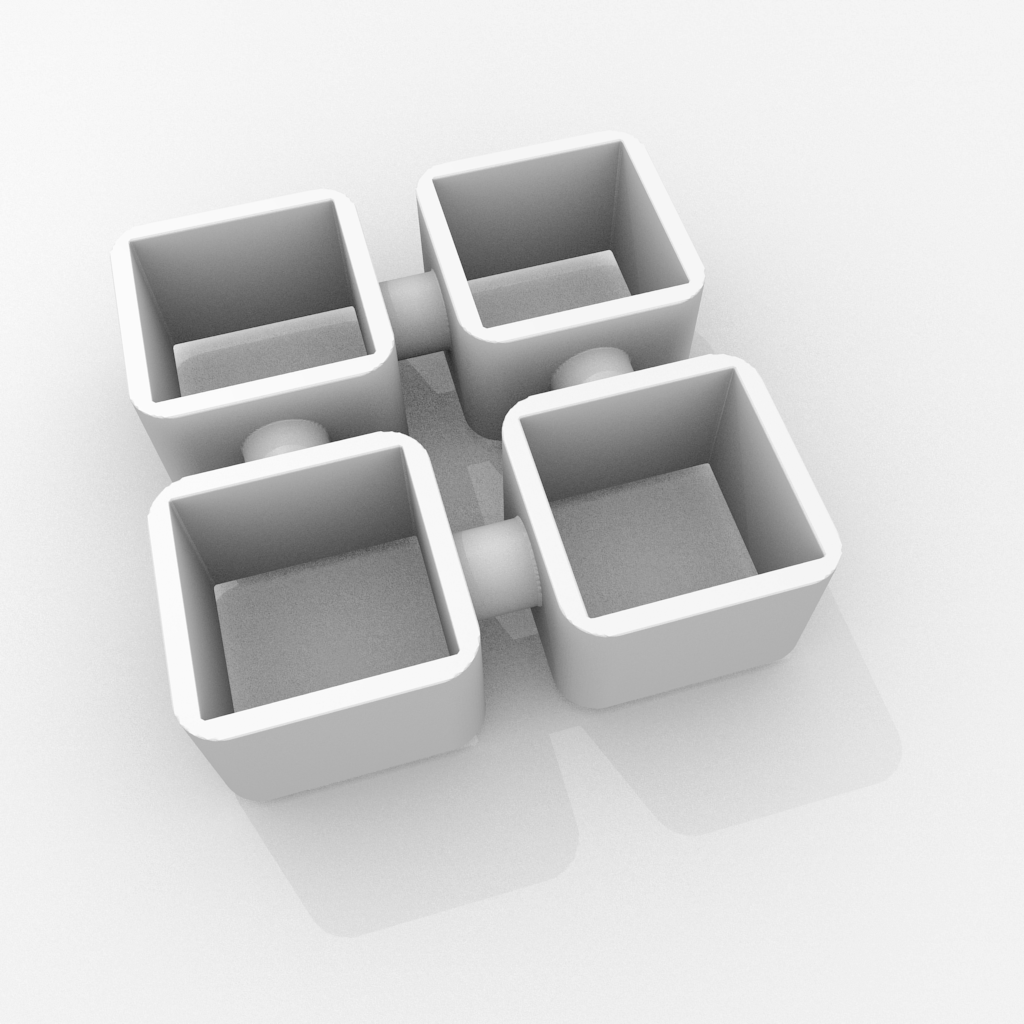}\label{fig:m7}}
	\subfigure[Model 8]{\includegraphics[width=25mm]{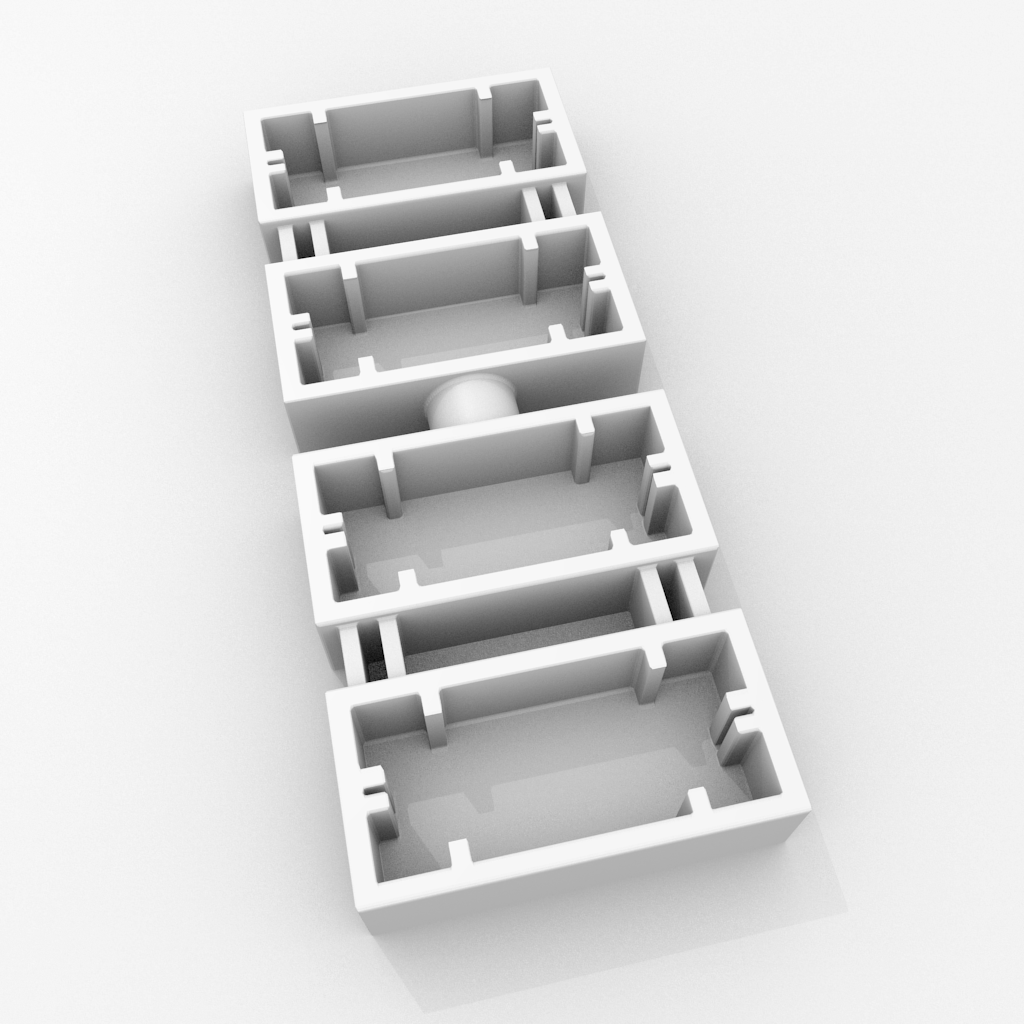}\label{fig:m8}}		
        \subfigure[Model 9]{\includegraphics[width=25mm]{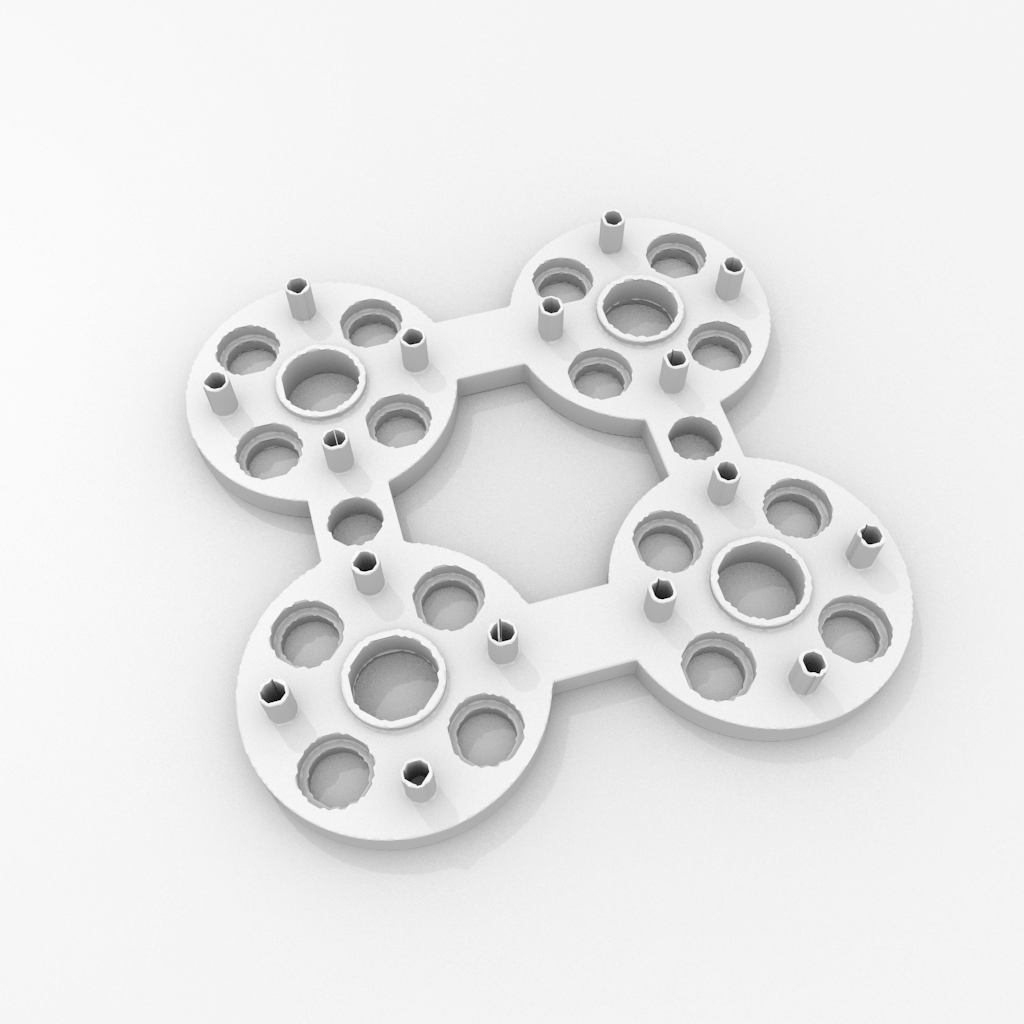}\label{fig:m9}}
	\subfigure[Model 10]{\includegraphics[width=25mm]{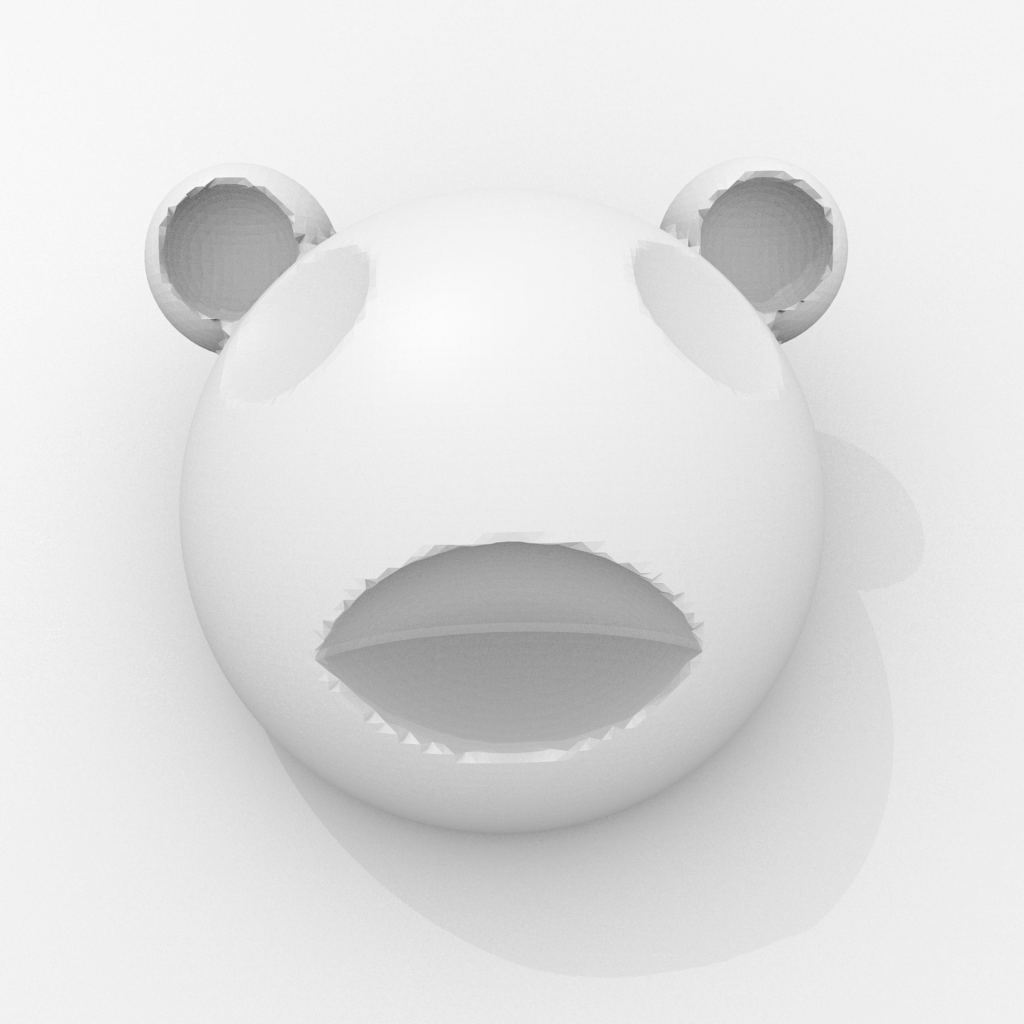}\label{fig:m10}}	
	\subfigure[Model 11]{\includegraphics[width=25mm]{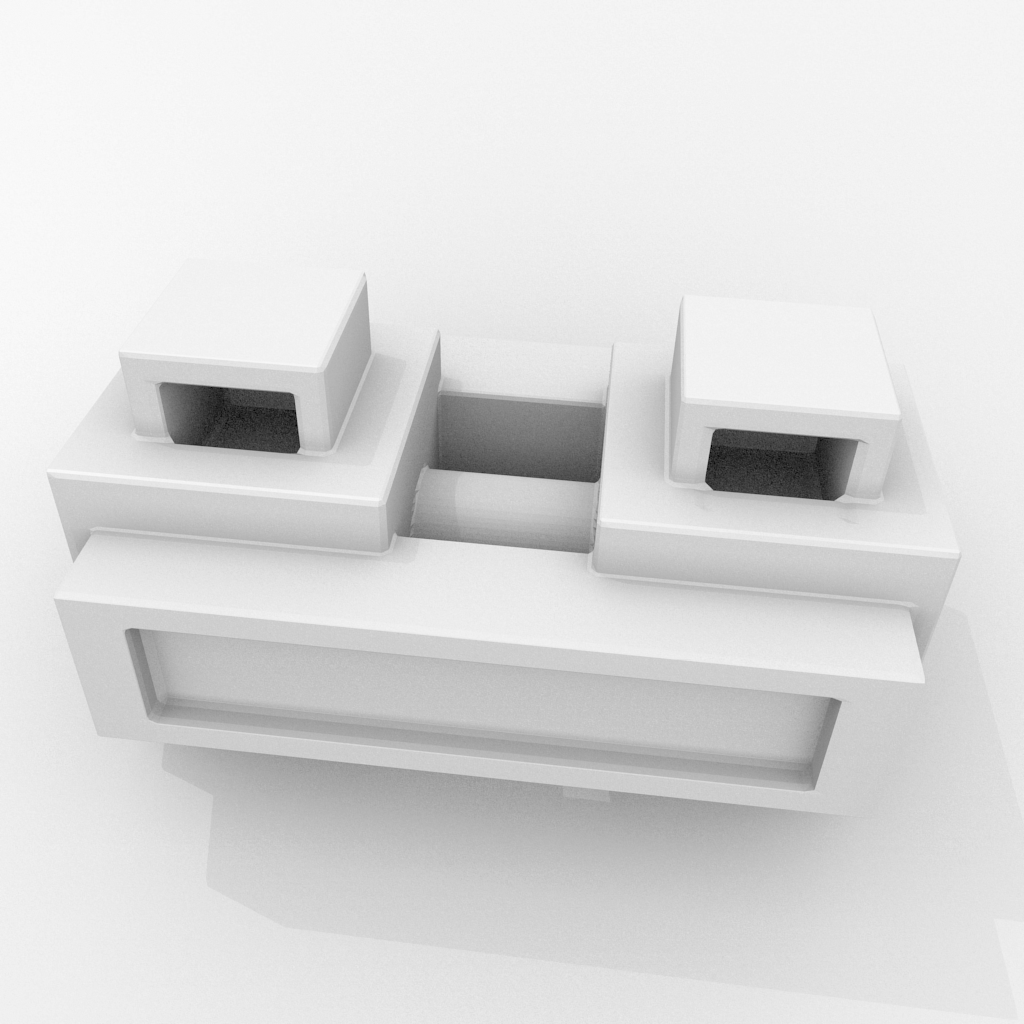}\label{fig:m11}}	
	\caption{Models used for the evaluation.\label{fig:model}}
\end{figure*}

\begin{figure*}[!htbp]
	\centering
	\includegraphics[width=160mm]{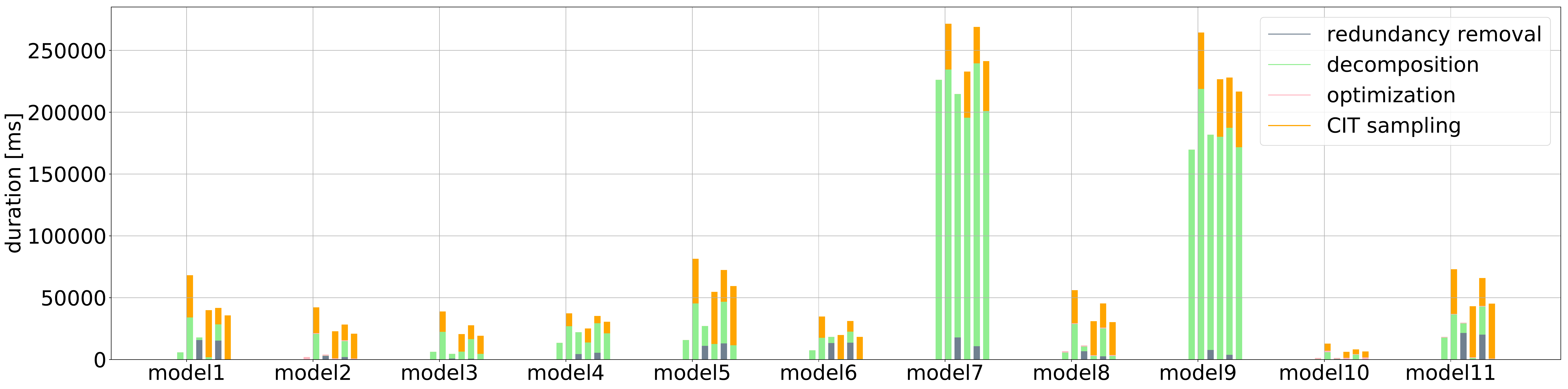}
	\caption{Timings for different pipeline configurations using data set $1$. For each model configurations from left to right: $(0,0,0)$, $(1,0,0)$, $(0,0,1)$, $(0,1,1)$, $(1,0,1)$ and $(1,1,1)$.} 
	\label{fig:pipeline_timings_1}
\end{figure*}

\begin{figure*}[!htbp]
	\centering
	\includegraphics[width=160mm]{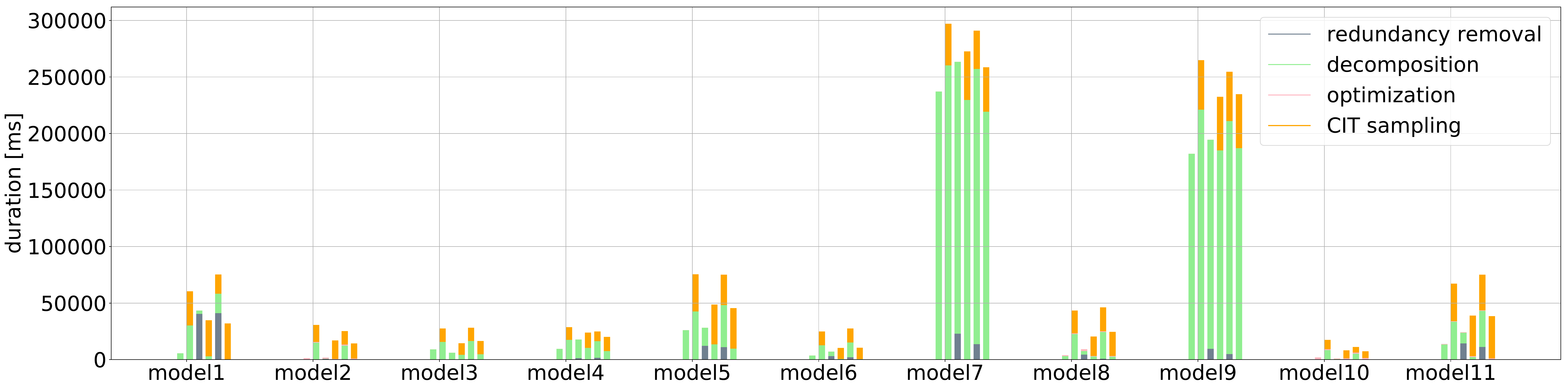}
	\caption{Timings for different pipeline configurations using data set $2$. For each model configurations from left to right: $(0,0,0)$, $(1,0,0)$, $(0,0,1)$, $(0,1,1)$, $(1,0,1)$ and $(1,1,1)$.} 
	\label{fig:pipeline_timings_2}
\end{figure*}


\begin{figure*}[htb]
	\centering
	\subfigure[Data set $1$]{\includegraphics[width=160mm]{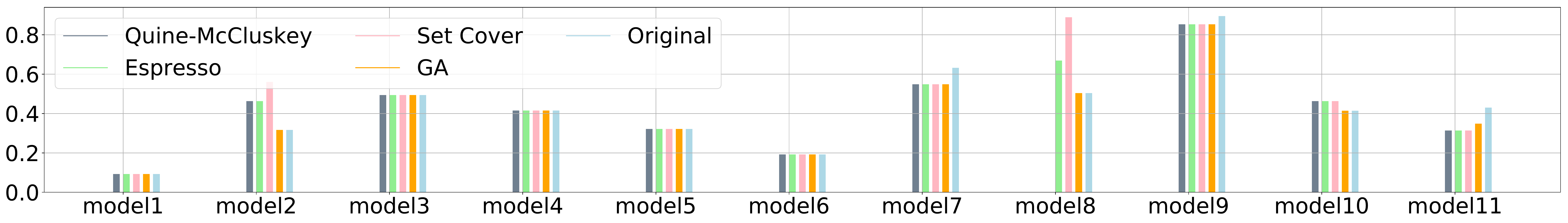}\label{fig:opt_size_1}}
	\\
	\subfigure[Data set $2$]{\includegraphics[width=160mm]{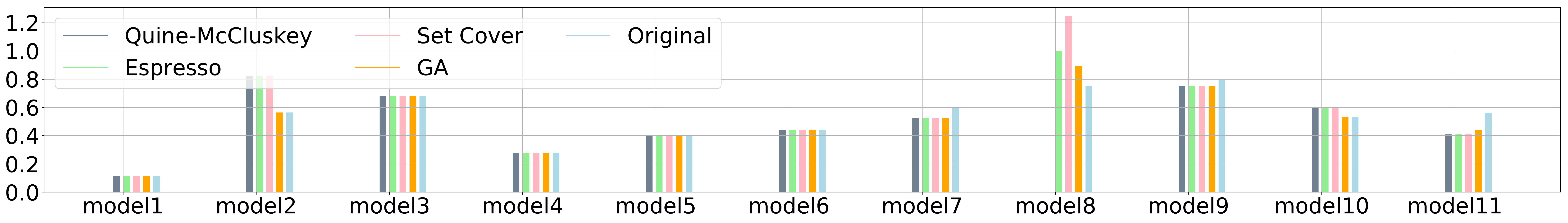}\label{fig:opt_size_2}}
	\caption{Ratio of input and output tree size for both data sets. The light blue bar 'Original' indicates the size of the initial hand-crafted expression.}
	\label{fig:opt_size}
\end{figure*}

\begin{figure*}[htb]
	\centering
	\subfigure[Data set $1$]{\includegraphics[width=160mm]{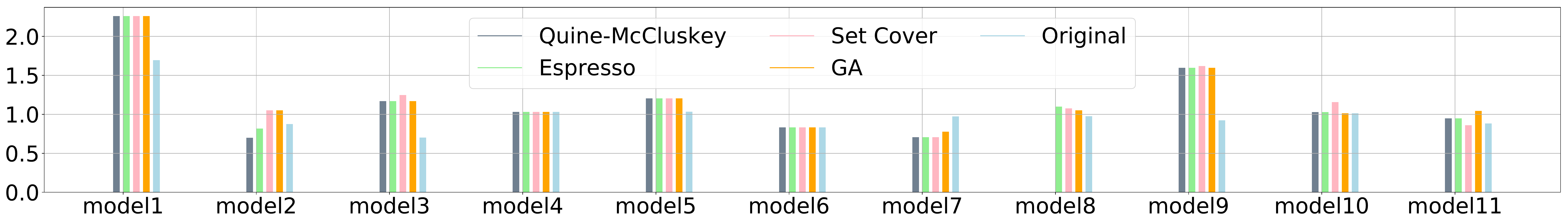}\label{fig:opt_prox_1}}
	\\
	\subfigure[Data set $2$]{\includegraphics[width=160mm]{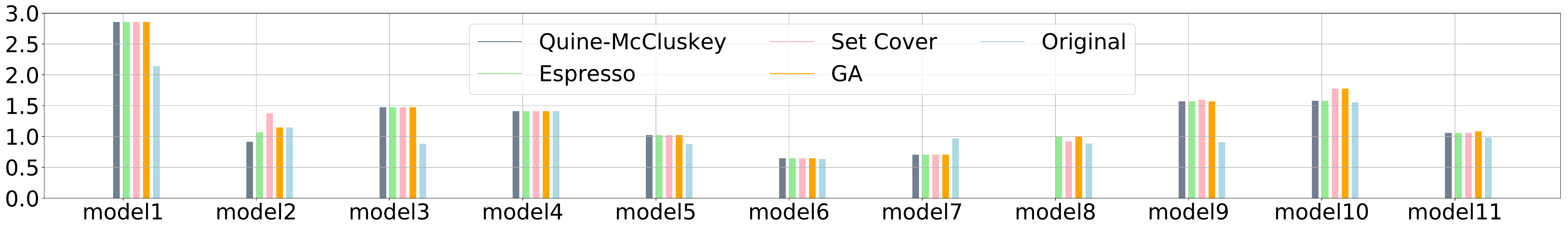}\label{fig:opt_prox_2}}
	\caption{Ratio of input and output tree proximity for both data sets. The light blue bar 'Original' indicates the proximity of the initial hand-crafted expression.}
	\label{fig:opt_prox}
\end{figure*}

\begin{figure*}[htb]
	\centering
	\subfigure[Model 2 (input and output trees are equal)]{\includegraphics[width=0.45\textwidth]{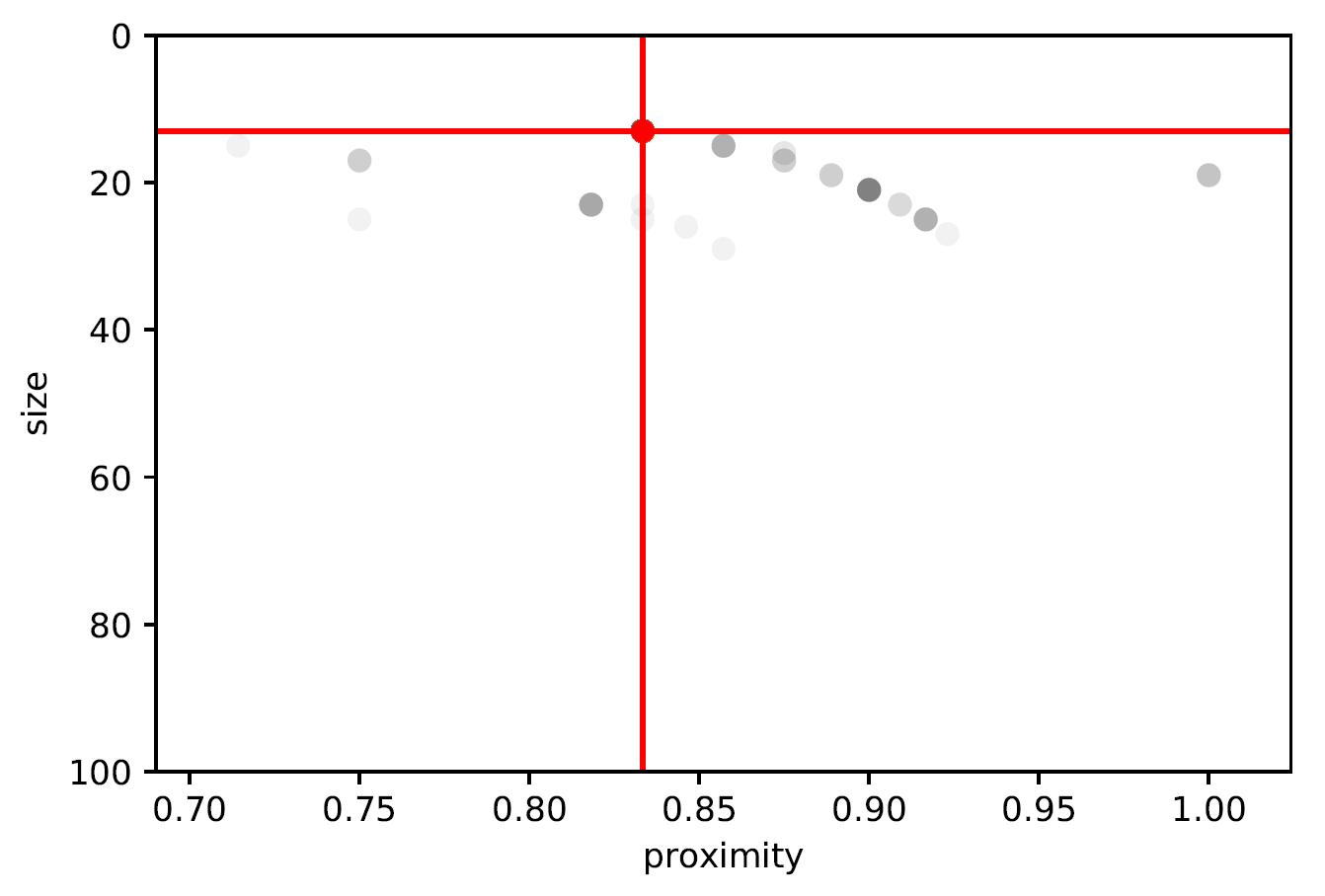}\label{fig:m2_ga}}
	\subfigure[Model 8]{\includegraphics[width=0.45\textwidth]{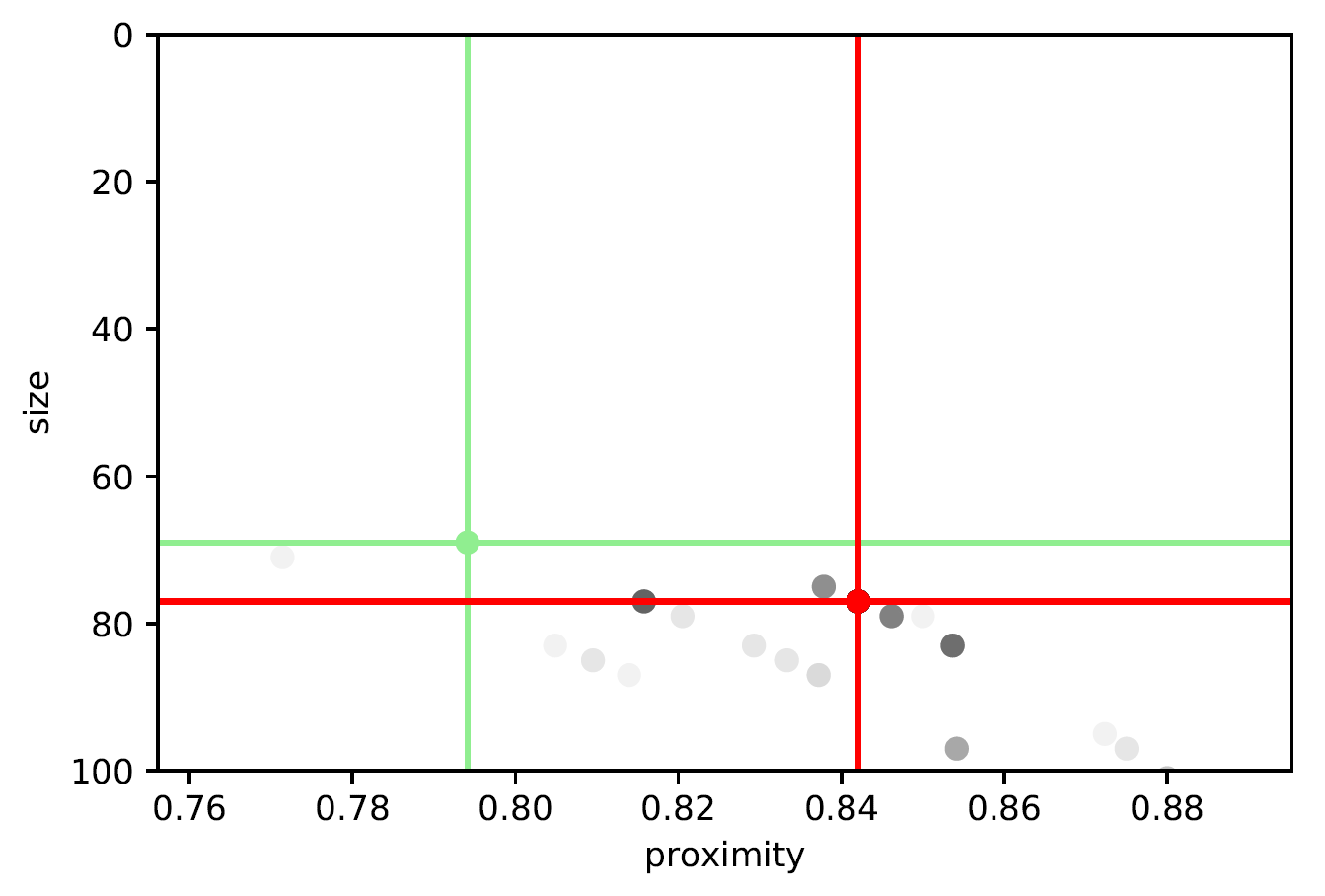}\label{fig:m8_ga}}
	\\
	\subfigure[Model 10]{\includegraphics[width=0.45\textwidth]{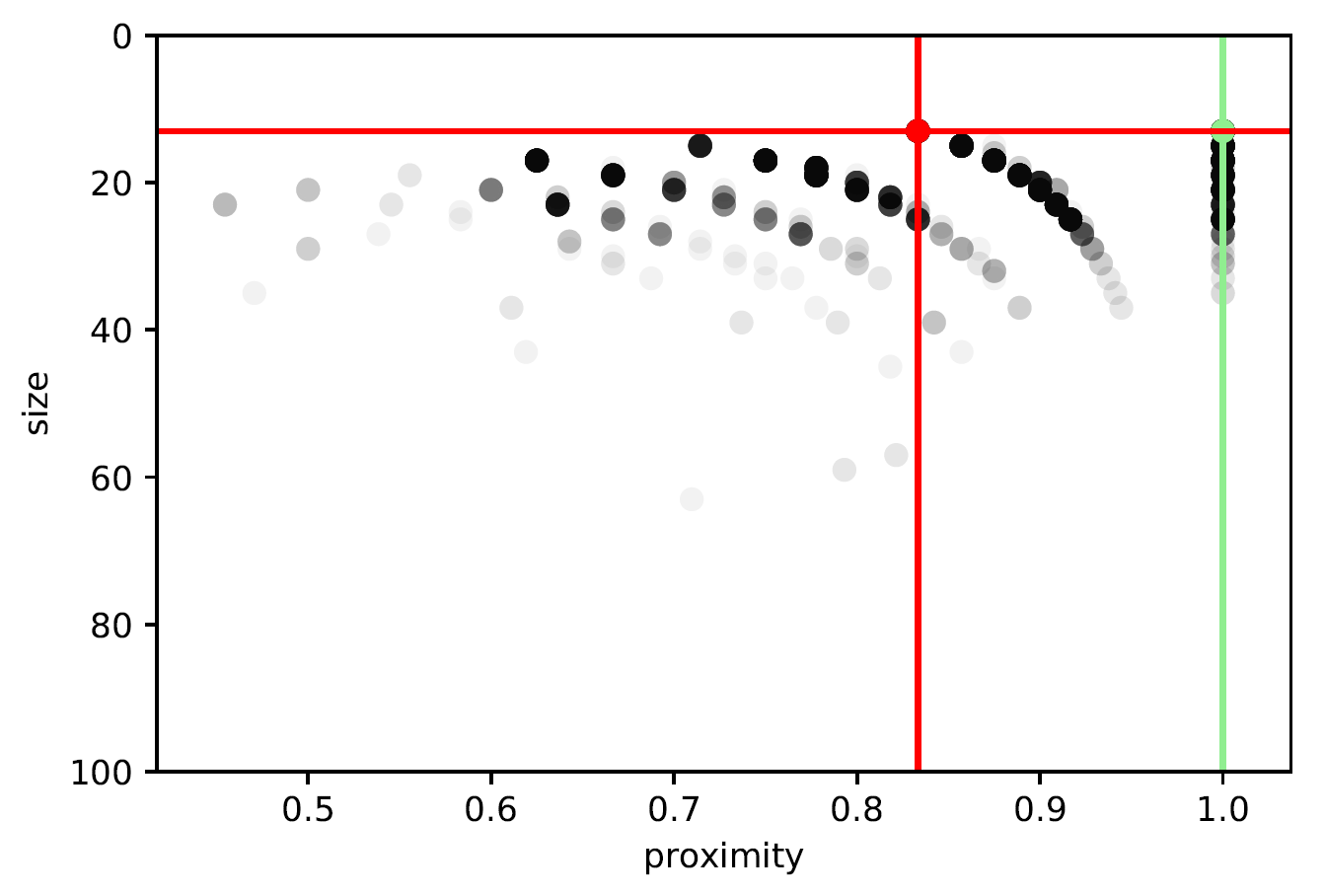}\label{fig:m10_ga}}
	\subfigure[Model 11]{\includegraphics[width=0.45\textwidth]{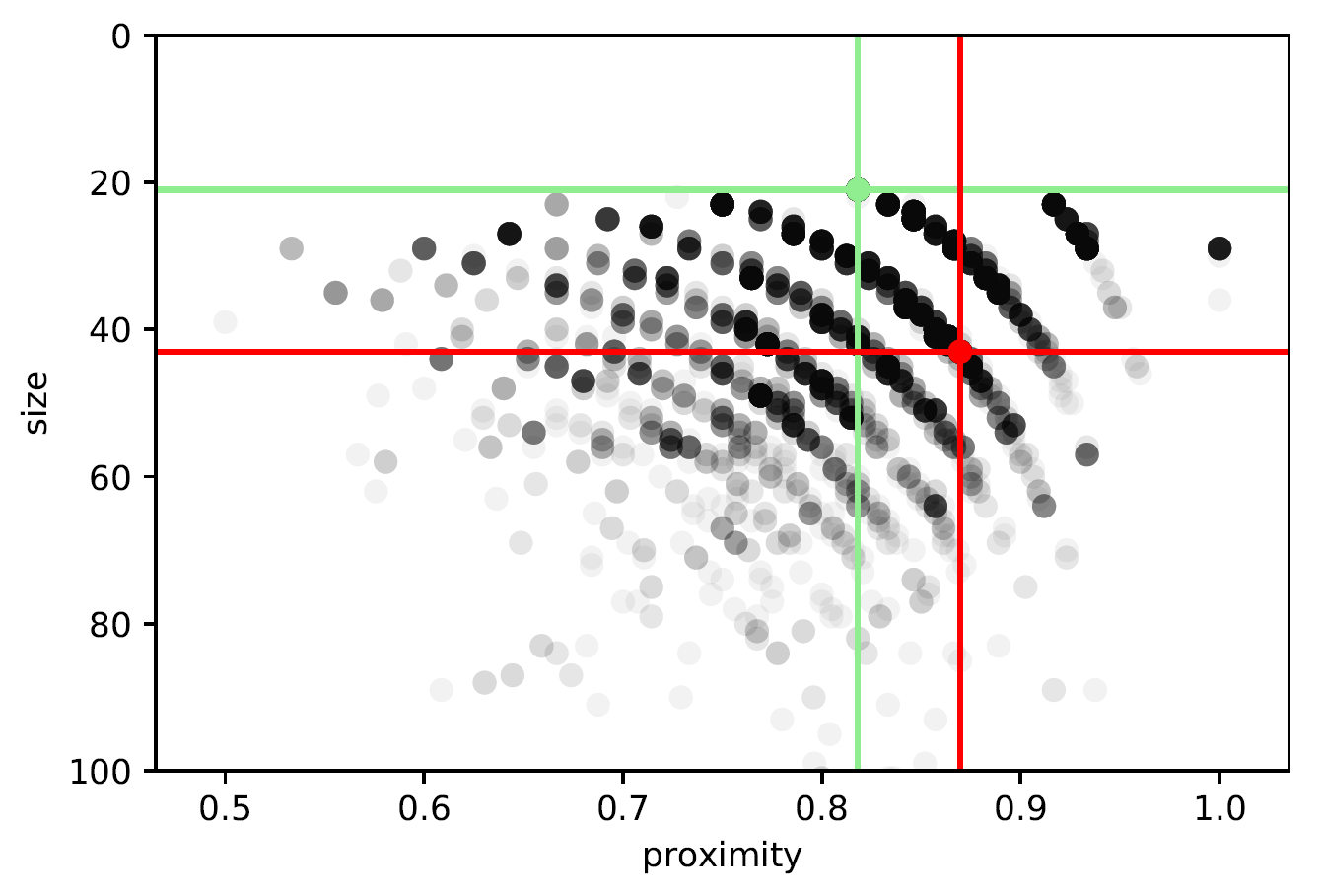}\label{fig:m11_ga}}
	\caption{All trees representing the remaining solid produced by the GA for data set $2$ with a geometric score of $1.0$. The darker the point, the more trees have exactly this size/proximity combination. The red dot indicates the input tree that represents the remaining solid, the green one the selected resulting tree.}
	\label{fig:ga_models}
\end{figure*}